%% file: main.tex
\documentclass{bmvc2k}
\input{preamble.tex}

\makeatletter
\let\BMVA@blfootnote\@gobble   
\makeatother

\title{%
 Crane: \underline{C}ontext-Guided P\underline{r}ompt Learning and \underline{A}ttention Refi\underline{ne}ment for Zero-Shot Anomaly Detection
}

\addauthor{Alireza Salehi$^*$}{alireza.salehi@postgrad.manchester.ac.uk}{3,1}
\addauthor{Mohammadreza Salehi}{s.salehidehnavi@uva.nl}{2}
\addauthor{Reshad Hosseini}{reshad.hosseini@ut.ac.ir}{3}
\addauthor{Cees G. M. Snoek}{cees.snoek@uva.nl}{2}
\addauthor{Makoto Yamada}{makoto.yamada@oist.jp}{5}
\addauthor{Mohammad Sabokrou}{m.sabokrou@newuu.uz}{4,5}

\addinstitution{
    University of Manchester\\
    Manchester, United Kingdom
}
\addinstitution{
    University of Amsterdam\\
    Amsterdam, Netherlands
}
\addinstitution{
    University of Tehran\\
    Tehran, Iran
}
\addinstitution{
    New Uzbekistan University\\
    Tashkent, Uzbekistan
} 
\addinstitution{
    Okinawa Institute of Science and Technology\\
    Okinawa, Japan
}
\runninghead{Salehi et al.}{Crane for Zero-shot Anomaly Detection}

\newcommand{\methodname}{\texttt{Crane} }
\newcommand{\methodnamep}{\texttt{Crane}$^+$ }
\newcommand{\methodnamepp}{(\texttt{Crane}$^+$) }
\newcommand{\methodnamepd}{\texttt{Crane}$^+$.}
\newcommand{\methodnamepc}{\texttt{Crane}$^+$, }

\newif\ifusecolor
\usecolorfalse
\newcommand{\mytextcolor}[2]{\ifusecolor\textcolor{#1}{#2}\else#2\fi}

\begin{document}

\maketitle

\begingroup
\renewcommand{\thefootnote}{*}
\footnotetext{This work was mainly conducted during author's internship at the Okinawa Institute of Science and Technology.
\vspace{2em}}
\endgroup

\newcommand{\clipattn}{E-Attn\xspace}
\newcommand{\dinoattn}{D-Attn\xspace}

\input{sec/0_abstract_3}
\clearpage
\input{sec/1_introduction_1}

\input{sec/2-ralated_works}
\input{sec/3-method_wo_image_index}
\input{sec/4-experiments_1}


\bibliography{references}

\input{sec/5-appendix_1}

\end{document}

%% file: preamble.tex
\usepackage{makecell}
\usepackage{multicol}

\usepackage{microtype}
\usepackage{graphicx}
\usepackage{booktabs} 
\usepackage{pifont}
\usepackage{booktabs}
\usepackage{multirow}
\usepackage{array}
\usepackage{adjustbox} 
\usepackage[table]{xcolor} 
\usepackage{tikz}
\usepackage{placeins} 
\usepackage{float}

\usepackage{xspace}

\usepackage{amsmath}
\usepackage{amssymb}
\usepackage{mathtools}
\usepackage{amsthm}

\usepackage{capt-of}
\usepackage{hyperref}
\usepackage{url}

%% file: sec/0_abstract_3.tex
\begin{abstract}
Zero-shot anomaly detection and localization aims to learn from source-domain data and generalize to unseen target domains without target-domain samples. Recent CLIP-based methods perform inference by comparing visual features with normal and abnormal textual prototypes; however, dense localization remains substantially weaker than image-level detection, especially for small or subtle abnormal regions. We identify two key limitations behind this gap: CLIP's vision encoder is primarily optimized for global image--text alignment, which limits its ability to preserve fine-grained spatial details, and the learned normal/abnormal text representations are not sufficiently aligned with dense visual features. To address these limitations, we propose \methodname (\underline{C}ontext-Guided P\underline{r}ompt Learning and \underline{A}ttention Refi\underline{ne}ment), a CLIP-based framework that first adapts the vision encoder with a correlation-based attention module to better preserve local visual structure. Second, it conditions learnable normal and abnormal prompts on global image context during training, improving instance-aware text--vision alignment. Third, it incorporates anomaly-aware local-to-global fusion, which injects anomaly-relevant patch features into the global image representation for more sensitive image-level detection. Finally, we show that the same correlation-based attention design can exploit spatial correlations from DINOv2, yielding a boosted variant, \methodnamepc with stronger localization ability. 
Across seven industrial benchmarks, \methodname improves mean image-level AP by 4.5\% over the strongest compared baseline, while \methodnamep improves mean pixel-level AUPRO by 9.0\%. The code is available at \url{https://github.com/Alireza99Salehi/Crane}.
\vspace{1em}
\end{abstract}

%% file: sec/1_introduction_1.tex
\section{Introduction}
\vspace{-0.5em}
Image anomaly detection aims to identify images, or regions within images, that deviate from normal visual patterns~\cite{salehi2021unified}. This task is particularly important in practical settings where abnormal samples are rare, diverse, and difficult to collect for each target domain, while normal data is often more abundant. For example, in medical diagnostics, datasets contain many healthy patient scans but few from those with rare conditions~\cite{salehi2021unified}. Similarly, in industrial defect detection~\cite{liu2024deep} and self-driving cars, normal data is abundant, but anomalies such as manufacturing defects or road obstacles are rare yet critical to detect~\cite{bogdoll2022anomaly}. In these cases, a model is trained on normal data to identify abnormalities.

\input{imgs/first_page_prelude/fig_1}

Recent studies~\cite{zhou2023anomalyclip, cao2024adaclip, jeong2023winclip, vcpclip} show that collecting normal data from all target domains is often impractical, due to factors such as privacy concerns. To address this, zero-shot anomaly detection is introduced, allowing models trained on source data to detect anomalies in unseen target datasets without requiring domain-specific samples. Current methods either leverage CLIP’s zero-shot capabilities with manually crafted prompts or adapt CLIP by learning prompts from source data to enable domain generalization. Despite progress, generalization remains inconsistent between image-level and pixel-level performance. For instance, as shown by Figure~\ref{fig:overall_results}, two state-of-the-art methods, AdaCLIP~\cite{cao2024adaclip} and AA-CLIP~\cite{ma2025aa}, get high image-level results while significantly compromising pixel-level performance (50.1\% and 42.0\% AUPRO) on industrial benchmarks, highlighting challenges in anomaly localization.

We attribute this gap to two key challenges: (1) Abnormal regions are often small, and the coarse-grained pretraining of CLIP limits the sensitivity of its global features to anomalous patterns, reducing its ability to distinguish normal samples from abnormal variations. (2) CLIP's dense features are not well-suited for segmentation tasks due to spatial misalignment~\cite{salehi2023time, wysoczanska2024clip, salehi2025mosic}, limiting the model’s ability to capture fine-grained details. Previous methods~\cite{zhou2023anomalyclip} address these challenges by prompt learning or fine-tuning the vision encoder on the source data. However, in prompt learning, the text encoder often fails to generate representations that are discriminative enough for subtle anomalies due to its limited capacity and the coarse pretraining of the text encoder. Fine-tuning the vision encoder, as done in prior work~\cite{cao2024adaclip, lan2024clearclip}, can alleviate this issue in some domains, yet it can also limit generalization and lead to subpar performance in others, since the available training data is often limited and can encourage in-domain overfitting, ultimately hurting domain generalization.

To address these challenges more effectively, we propose \methodname (\underline{C}ontext-guided P\underline{r}ompt Learning and \underline{A}ttention Refi\underline{ne}ment). To learn more discriminative features for the text encoder, we guide prompts using the image classification token (CLS) from the image encoder, alongside other learnable parameters. This enables the model to generate representations conditioned on the image context, improving the modeling of fine-grained distributions in a data-efficient manner. Additionally, we introduce a local-to-global fusion mechanism that aggregates dense anomalous features into the global visual embedding, enhancing its sensitivity to local anomalous cues. 
To retain spatial alignment, we modify the CLIP vision encoder by introducing a correlation-based attention module that better captures fine-grained local information. Moreover, we introduce a simple method to inject spatial knowledge from vision encoders such as DINOv2~\cite{oquab2023dinov2} and DINOv3~\cite{simeoni2025dinov3} into our zero-shot framework--even though these encoders are not inherently zero-shot--further improving performance. We call this variant \methodnamepd 
~Our extensive evaluations across 14 datasets spanning medical
and industrial domains show that \methodname achieves strong performance relative to state-of-the-art approaches, with the most consistent gains observed on industrial benchmarks,
including dataset-averaged improvements of 0.7--4.5\% in anomaly detection and 0.2--5.7\% in localization,
while \methodnamep delivers additional average pixel-level gains of 0.3--3.3\% further demonstrating the method's
generalization ability for complex and fine-grained domains.

%% file: imgs/first_page_prelude/fig_1.tex

\begin{figure}[!t]
    \centering
    \setlength{\tabcolsep}{0pt}

    \begin{tabular}{ccc}
        \begin{minipage}{0.55\textwidth}
            \centering
            \includegraphics[width=\textwidth]{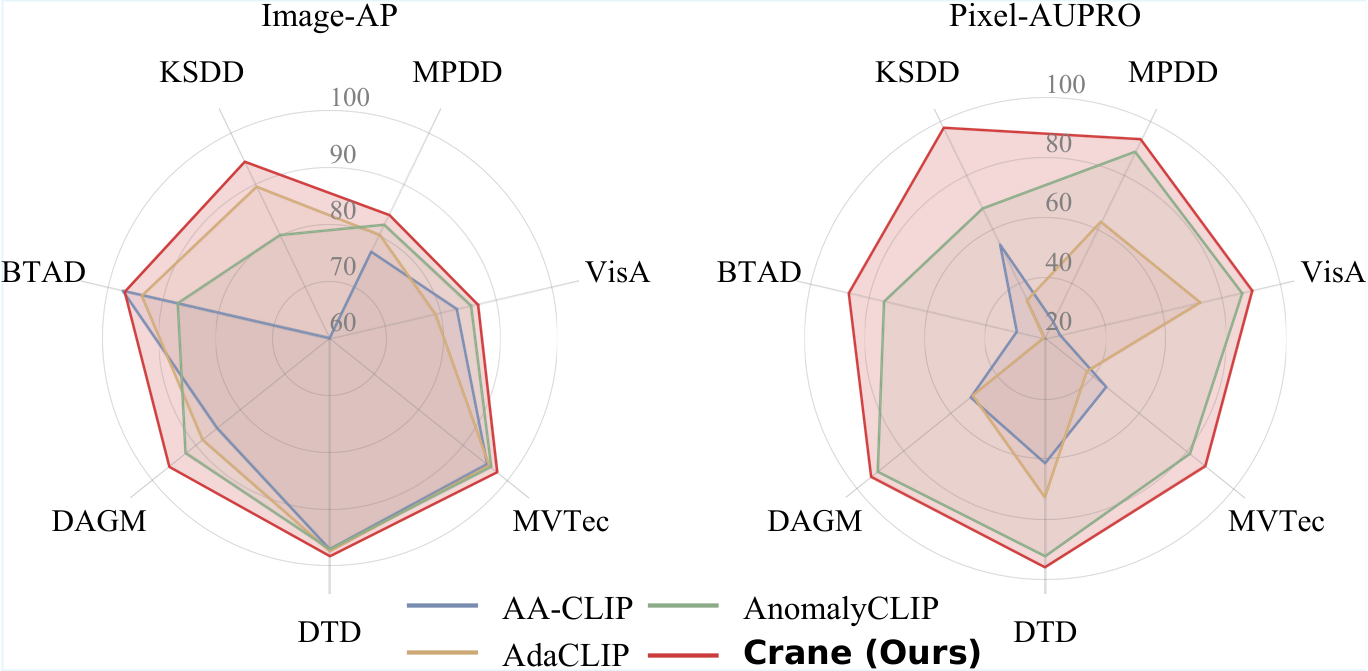} \\[-6pt]
        \end{minipage}
        & 
        \begin{minipage}{0.01\textwidth}
            \centering
            \begin{tikzpicture}
                \draw[dashed, thick, gray] (0, -0.3) -- (0, 3);
            \end{tikzpicture} \\
        \end{minipage}
        & 
        \begin{minipage}{0.44\textwidth}
            \centering
            \includegraphics[width=\textwidth]{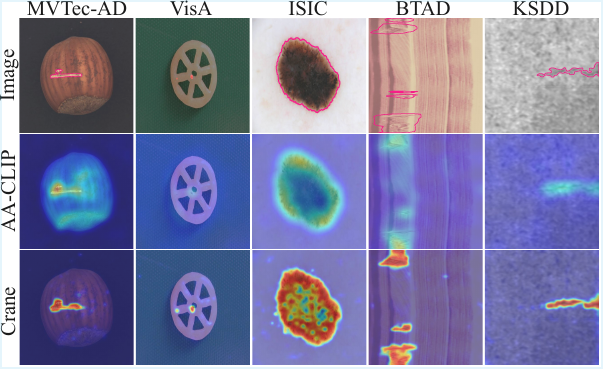} \\
        \end{minipage}
    \end{tabular}

    \vspace{1em}
    \caption{\textbf{Zero-shot anomaly detection performance of \methodname compared to state-of-the-art methods.} The radar plots summarize image and pixel level performances, showing \methodname consistently improves performance across industrial benchmarks. Qualitative examples further demonstrate that \methodname produces more precise localizations than competing approaches. Anomalous regions outlined in pink.} 
    \label{fig:overall_results}
    \vspace{-.6em}
\end{figure}

%% file: sec/2-ralated_works.tex
\vspace{-0.8em}
\section{Related Works}
\vspace{-.3em}
\paragraph{Unsupervised \& Semi-supervised Anomaly Detection} 
Unsupervised and semi-supervised anomaly detection are dominantly used methods in the field; their main assumption is there is access to enough normal samples from the target domains. For unsupervised anomaly detection, a common strategy leverages pretrained models~\cite{salehi2021unified, liu2024deep} to extract discriminative features, modeling the normal distribution through mechanisms 
like knowledge distillation~\cite{mkd, rd4ad, efficientad}, memory banks~\cite{patchcore, memkd}, reconstruction-based methods~\cite{fastrecon}, 
and flow-based techniques~\cite{fastflow, cflowad}. As the anomaly data is unavailable during training, some methods generate synthetic anomalies using self-supervised methods~\cite{simplenet, generalad}, data augmentation~\cite{zou2022spot-visa}, or generative models~\cite{mirzaei2023fake, hu2024anomalydiffusion, chen2024unified}. Some works directly use the diffusion models to better model normal data, resulting in better detection~\cite{yao2024glad, fuvcka2024transfusion}. Semi-supervised anomaly detection methods incorporate a few anomalous samples during training~\cite{ruff2019deep, pang2019deep, bgad, prn, dra} to cope with the lack of abnormal samples during training. Although effective, these methods assume the availability of enough normal samples from the target domain, differing from our objective. In contrast, we evaluate generalization performance on a target dataset while exclusively training on source data independent of that target, which is explained in the zero-shot anomaly detection.  

\vspace{-.5em}
\paragraph{Zero-shot anomaly detection}
Zero-shot anomaly detection methods assume no access to the target dataset; instead, they leverage foundation models~\cite{radford2021learning,blip,sam}, pretrained on large-scale datasets, to learn generalizable features from source data that can be applied to unseen target datasets. Some works explore language-free ZSAD by learning normal and abnormal representations directly in the visual feature space using pretrained vision foundation models~\cite{hou2026visualad,fuvcka2026anomalyvfm}. Most existing works, though, build on contrastive vision-language models such as CLIP~\cite{radford2021learning}, which aligns global visual embeddings with textual descriptions. However, CLIP struggles with patch-level misalignment and lacks domain-specific sensitivity, limiting its ability to detect fine-grained anomalies. To address this issue, early methods focused on designing manually curated prompt templates~\cite{jeong2023winclip,deng2023anovl,chen2023april,ssa}, which depend on domain knowledge and prompt quality. 
Whereas more recent works adopt prompt learning techniques~\cite{coop, bayesian_prompt, coprompt} to automate prompt optimization in a few-shot setting. For instance, 
AnomalyCLIP~\cite{zhou2023anomalyclip} introduced object-agnostic prompts, simplifying prompt crafting while utilizing general anomalous patterns. 
To address the patch-level misalignment, some methods fine-tune the vision encoder~\cite{vcpclip,clipad,clipsam,inctrl, chen2023april}, keep the vision encoder frozen yet further refine its attention modules~\cite{promptad,zhou2023anomalyclip}, or use deep token tuning in both text and vision encoder~\cite{cao2024adaclip}, or even substitute CLIP entirely with a more recent spatially-aware contrastive VLM such as Tipsomaly~\cite{tipsomaly}. AnomalyCLIP~\cite{zhou2023anomalyclip} follows the second approach, using extra ``VV'' attention introduced in CLIPSurgery~\cite{clipsurgery} to leverage patch embedding correlations, enhancing CLIP vision encoder segmentation ability. AA-CLIP~\cite{ma2025aa} and AdaCLIP~\cite{cao2024adaclip} follow the latter approach by jointly tuning the vision and text encoders. AdaCLIP enhances feature alignment by applying k-means clustering to dense visual features and adding learnable linear projection heads to the vision encoder, while AA-CLIP performs step-by-step fine-tuning of the text and vision encoders. Although effective, our experiments reveal that AnomalyCLIP struggles with image-level generalization, whereas AdaCLIP and AA-CLIP underperform at pixel-level detection. To address these shortcomings, we propose a context-guided prompt learning strategy to enhance alignment between textual and visual features and
extend the technique proposed by CLIPSurgery. \mytextcolor{orange}{We further improve spatial alignment using models such as DINOv2 motivated by \cite{shi2024harnessing, lan2024proxyclip}}. Unlike AdaCLIP, we do not fine-tune the vision encoder, as this can degrade its performance~\cite{zhai2022lit}, and we avoid clustering techniques such as K-Means, which require additional hyperparameter tuning. 

%% file: sec/3-method_wo_image_index.tex
\vspace{-1.2em}
\section*{Problem Statement}
\vspace{-.5em}
Let \( M_{\theta} \) denote a pretrained vision-language model (e.g., CLIP) with fixed parameters \( \theta \). We consider source anomaly detection datasets \( D_{\text{train}} \) from selected domains, where 
each image \( x \in D_{\text{train}} \subset \mathbb{R}^{C \times H \times W} \) is paired with image-level and pixel-level labels
\[
y \in \{0,1\}, \quad S \in \{0,1\}^{H \times W},
\]
where \( y = 1 \) indicating an anomaly and \( y = 0 \) a normal sample and pixels labeled as 1 marking anomalous regions.

In zero-shot anomaly detection framework, given a prompt \( P \)
the model produces two continuous anomaly scores for each image \( x \):
\[
\hat{y},\, \hat{S} = M_{\theta}(x, P),
\]
where \( \hat{y} \in [0,1] \) is the image-level anomaly score and \( \hat{S} \in [0,1]^{H \times W} \) is the pixel-level anomaly map. Here, \( P \) comprises textual templates (e.g., ``a photo of normal CLS'') optionally augmented with learnable parameters, which are either prepended to or integrated within the textual input. The final image-level decision—classifying an image as normal or abnormal—is then obtained by thresholding \( \hat{y} \) as follows:
\[
y' =
\begin{cases}
1, & \text{if } \hat{y} > \tau,\\[1mm]
0, & \text{otherwise},
\end{cases}
\]
where \( y' = 1 \) denotes an anomalous sample.

A common trend in zero-shot anomaly detection is to optimize the prompt \( P \) on \( D_{\text{train}} \) while keeping \( \theta \) fixed, so that \( P \) captures generalizable anomalous features. The optimized prompt \( P^{*} \) is then applied to new domains—where labeled anomaly data is unavailable—for both image- and pixel-level detection.

\section{Method} 
\begin{figure*}[t]    
    \centering
    \includegraphics[width=1\linewidth]{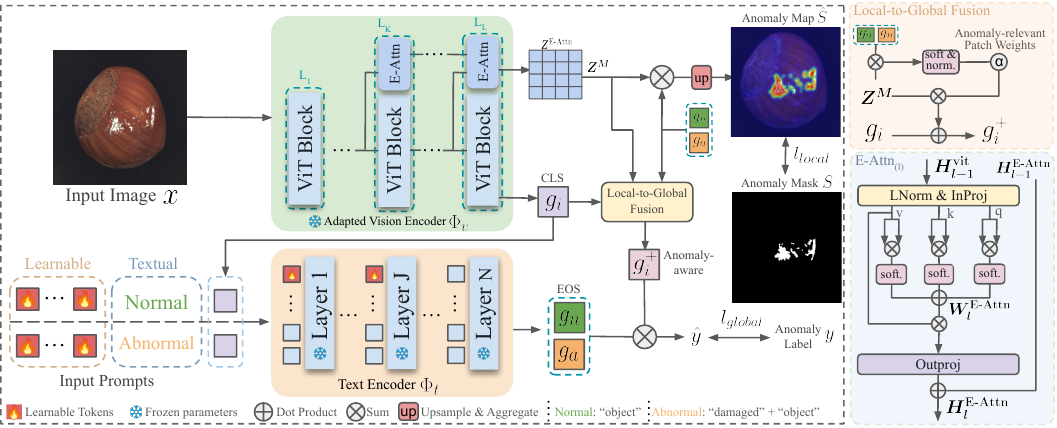}
    \vspace{-1.2em}
    \caption{
    \textbf{\methodname framework for Zero-shot Anomaly Detection.}
    First, we extract global embedding $g_i$ and spatially aligned local embeddings $Z^{\text{M}}$ by passing the image through the CLIP's adapted vision encoder $\Phi_v$. 
    Next, we guide learnable prompts with global image context to enhance the capture of fine-grained anomalous patterns. We then compute anomaly map $\hat{S}$ by measuring the similarity between visual embeddings and textual normal/anomalous embeddings. To boost image-level sensitivity to anomalies, we refine the global embedding by incorporating local embeddings weighted by their scores ($g_i^+$) and finally obtain anomaly score $\hat{y}$.}   
    \label{fig:framework}
\vspace{-.7em}
\end{figure*}

\vspace{-0.3em}
We propose a unified framework that leverages CLIP as a zero-shot backbone (\( M_{\theta} \)) for classification and segmentation, while adapting it for anomaly detection to bridge the domain gap between CLIP’s pretraining and specialized detection tasks. As shown in Figure~\ref{fig:framework}, we enhance the vision encoder with an \clipattn module that refines attention weights to better capture localized anomalous cues. We also learn class-agnostic input prompts (\( P \)) and trainable tokens in the text encoder (\(\Phi_t\)), guided by visual feedback from the vision encoder (\(\Phi_v\)). Additionally, an anomaly-aware local-to-global fusion mechanism integrates dense anomalous features into the global embedding, yielding more anomaly-sensitive representations for robust zero-shot generalization at both image and pixel levels across unseen domains.

\newcommand{\numVisEmbs}{N}     
\newcommand{\numTexEmbs}{T}     
\newcommand{\embdim}{D}     

\newcommand{\layeridx}{l}  
\newcommand{\layeridxnp}{l}     
\newcommand{\maxlayer}{L}  
\newcommand{\imageidx}{i}  

\vspace{-.6em}
\subsection{Local and Global Visual Feature Extraction}
\vspace{-.2em}

To craft the image anomaly score, $\hat{y}$, and the corresponding anomaly localization map, $\hat{S}$, in zero-shot, each input image needs to be modeled at both local (dense) and global (image-level) through our vision encoder $\Phi_v$ and then compared with the normal/abnormal text features extracted by the text encoder $\Phi_t$. Each component is explained in detail below.

\vspace{-1em}
\paragraph{Adapted CLIP Vision Encoder ($\Phi_v$).}
Given an input image $x$, the vision encoder produces a global embedding
$g_i \in \mathbb{R}^{\embdim}$, used for classification, and local embeddings
$Z^{\text{\clipattn}} \in \mathbb{R}^{\numVisEmbs \times \embdim}$, used for segmentation,
where $D$ denotes the embedding dimension and $N$ the number of patch embeddings
obtained from the E-Attn branch.
The global representation \( g_i \) is the [CLS] token from the forward pass of the CLIP vision transformer, which extracts textually aligned image-level features. However, for local embeddings, we opt not to use the ViT's original embeddings, as the emphasis on global image-text alignment during CLIP’s pretraining has led to degraded similarity between corresponding patch embeddings across layers, resulting in inaccurate segmentation~\cite{clipsurgery, lan2024clearclip}.

To address this, we adapt CLIP by replacing the Query-Key-based ($QK^{T}$) attention weighting with a self-correlation weighting scheme to reinforce semantic correlation across layers. Given $E \in \mathbb{R}^{\numVisEmbs \times \embdim}$ as a set of $N$ embeddings, a self-correlation attention weighting can be defined as:
\[
A(E) = \operatorname{softmax}\Bigl(\frac{EE^\top}{\sqrt{\embdim}}\Bigr), \quad A(E) \in \mathbb{R}^{\numVisEmbs \times \numVisEmbs},
\]
where $A(E)$ is a weight matrix that captures pairwise similarities between embeddings. We extend self-correlation weighting by applying the \textit{Extended Self-Correlation Attention (\clipattn)} branch at layer $\layeridxnp$ of CLIP ViT. Given $K_{\layeridx}, Q_{\layeridx}, V_{\layeridx} \in \mathbb{R}^{\numVisEmbs \times \embdim}$ as the \textit{key}, \textit{query}, and \textit{value}, we compute attention weights as follows:
\[
W^{\text{\clipattn}}_{\layeridx} = A\bigl(K_{\layeridx}\bigr) + A\bigl(Q_{\layeridx}\bigr) + A\bigl(V_{\layeridx}\bigr),
\]
Then, the same as the standard attention~\cite{dosovitskiy2020image}, $W^{\text{\clipattn}}_{\layeridx}$ is used to aggregate $V_{\layeridx}$ tokens, producing $H^{\text{\clipattn}}_{\layeridx}$ as the attention output. 
Since early layers have less expressive representations, this branch operates on the last \( K \) layers of the ViT. 
Given \( \maxlayer \) as the total number of layers, for input image $x$ the final
\clipattn output is obtained by aggregating intermediate outputs:
\[
Z^{\text{\clipattn}} =
\sum_{l=L-K+1}^{L} H_l^{\text{\clipattn}},
\qquad
Z^{\text{\clipattn}} \in \mathbb{R}^{\numVisEmbs \times \embdim}.
\]

\vspace{-1.5em}
\paragraph{Text Encoder ($\Phi_t$).} To obtain the optimum textual alignment with anomalous visual embeddings, we employ prompt learning and deep token tuning within the text encoder. The standard transformer block $t_{\layeridx}(.)$ at layer $l$ is defined as:
\[
[\operatorname{\text{\small{BOS}}}_{\layeridx},H_{\layeridx},\operatorname{\text{\small{EOS}}}_{\layeridx}] = t_{\layeridx}\bigl([\operatorname{\text{\small{BOS}}}_{(\layeridxnp-1)},H_{(\layeridxnp-1)},\operatorname{\text{\small{EOS}}}_{(\layeridxnp-1)}]\bigr)
\]
where [$\operatorname{BOS}_{\layeridx}$] and [$\operatorname{EOS}_{\layeridx}$] are special tokens marking the sequence’s start and end, and $H_{\layeridx}$ represents intermediate token representations. For each layer $l=2..J+1$, we replace the first $M$ tokens of $H_{(l-1)}$ with learnable tokens $\tau_{\layeridx} \in \mathbb{R}^{M \times \embdim}$ to capture anomaly-specific knowledge. At the final layer, the [$\operatorname{EOS}$] token, which aggregates the semantic representation of the input prompt, is used as the textual feature.

To design the input prompt, we abandon class-based, manually crafted templates in favor of object-agnostic learnable prompts~\cite{zhou2023anomalyclip}. As a result, we learn only two prompts (normal and anomalous), instead of two per dataset category. This approach leverages the shared structural patterns of anomalies across domains and reduces the need for domain-specific prompt engineering.
To achieve this, we use a set of $E$ learnable tokens for each of the normal and anomalous prompts, denoted by $\tau^n, \tau^a \in \mathbb{R}^{E \times \embdim}$, and concatenate them with general product and state textual descriptions: “object” for the normal case and “damaged object” for the anomalous case, appended after the learnable sets.

Additionally, we introduce \textit{context-guided prompt learning}, which integrates the image-level representation \( g_i \) into textual prompts during training, enabling the model to better capture fine-grained distributions.
In summary, the normal $g_n$ and abnormal textual embedding $g_a$ are constructed as:
\vspace{-1em}
\[
\begin{aligned}
g_n &= \Phi_t\Bigl([\tau^n_1, \tau^n_2, \dots, \tau^n_E, \text{``object''}, g_i]\Bigr), \\
g_a &= \Phi_t\Bigl([\tau^a_1, \tau^a_2, \dots, \tau^a_E, \text{``damaged''}, \text{``object''}, g_i]\Bigr).
\end{aligned}
\]

\paragraph{Calculating anomaly likelihood.}
Having obtained the textual \( G = \{g_n, g_a\} \) and visual embeddings \( e \in \{ g_i, Z_{j,k} \} \), where \( Z_{j,k} \) represents patch embedding at the position \( (j,k) \) of unflattened local branch $Z \in Z^M = Z_{Crane} = \{Z^{\text{E-Attn}}\},
$
or
$ Z \in Z^M = Z_{Crane^+} = \{Z^{\text{E-Attn}},Z^{\text{D-Attn}}\}$. 
for input image $x$, we can now compute the likelihood of each visual embedding belonging to 
the anomalous class (\( p_a \)) by applying the Softmax function to cosine similarity scores:
\begin{equation}
    p_a(e, G) = \frac{\exp\left( \langle e, g_a \rangle / T \right)}{\exp\left( \langle e, g_a \rangle / T \right)+\exp\left( \langle e, g_n \rangle / T \right)}
    \label{eq:likelihood}
\end{equation}
where the temperature \( T \) is set to 0.01 according to CLIP's hyperparameter details~\cite{radford2021learning}. We denote the probability of a visual embedding $e$ being abnormal, $p_a(e, G)$, as its anomaly score.

\vspace{-1em}
\paragraph{Anomaly-aware Local-to-Global Fusion} 
The image-level representation $g_i$, trained to capture a global representation of an image, may fail to encode the discriminative fine-grained anomalous cues due to its global focus. 
To address this issue, we propose a \textit{score-based spatial pooling} mechanism that fuses patch-level features into \( g_i \) based on their anomaly scores, ensuring the final representation preserves global semantics while capturing anomaly cues. For each local feature map \( Z \in \mathbb{R}^{H \times W \times \embdim} \) from $Z^M$, The anomaly-aware representation is constructed as:
\begin{equation*}
    g^z_a = \frac{\sum_{j,k} p_a(Z_{j,k}, G) Z_{j,k}}{\sum_{j,k} p_a(Z_{j,k}, G)}, 
    \quad Z \in Z^M.
\end{equation*}
The anomaly-aware representations from the available local branches are then aggregated as:
\[
\bar{g}_a =
\frac{1}{|Z^M|}
\sum_{Z \in Z^M} g_a^Z,
\qquad
g_i^+ = \frac{g_i+\bar{g}_a}{2},
\]
where $g_i^+$ is the anomaly-aware global representation used for
anomaly classification.

\vspace{-1em}
\paragraph{\mytextcolor{orange}{\methodnamep for enhanced localization.}}%
Domains such as medical imaging present unique challenges for anomaly detection, where anomalies often differ from normal tissue in texture or intensity by only a few pixels—requiring highly detailed feature discrimination. To address this, we propose a module that adapts DINOv2 to our zero-shot setting for more robust localization. While DINOv2 encodes spatially sensitive patch-level features, it lacks the semantic required for zero-shot capabilities. We integrate it through a module called \textit{DINO-guided spatial attention (\dinoattn)}, which—like \clipattn—replaces the standard attention weights with spatially aware ones, as described:
\begin{align*}
Z^{\text{D}} &= f_{\text{D}}(x), 
\quad S = \langle Z^{\text{D}}, Z^{\text{D}}\rangle, \\
W^{\text{\dinoattn}} &= \operatorname{softmax}(S + M), \quad
M_{ij} =
\begin{cases}
0, &  S_{ij} \ge \epsilon, \\
-\infty, &  S_{ij} < \epsilon,
\end{cases}
\end{align*}
where $Z^{\text{D}} \in \mathbb{R}^{\numVisEmbs \times \embdim}$ is DINOv2's output patch embeddings, the operator $\langle .,.\rangle$ is cosine similarity and $S \in \mathbb{R}^{\numVisEmbs \times \numVisEmbs}$ is the computed similarity matrix. To refine $S$, we apply a masking mechanism to discard low similarity scores, followed by softmax normalization over the last dimension. The refined similarity matrix $W^{\text{\dinoattn}}$ is then used at the final layer $\maxlayer$ as attention weights, producing attention output $H_{\maxlayer}^{\text{\dinoattn}} \in \mathbb{R}^{\numVisEmbs \times \embdim}$. In \methodnamepc \dinoattn's output $ Z^{\text{\dinoattn}} = H_{\maxlayer}^{\text{\dinoattn}}$  is added as a complementary local feature branch, yielding $Z^{\text{M}}$.  Notably, this module can be implemented efficiently, as DINOv2 features are extracted in parallel with CLIP. 

\subsection{Training}
To train the text encoder $\Phi_t$, we employ global and local loss functions. For input image $x$ the global loss ensures alignment between the global embedding $g_i^{+}$ and its corresponding textual embedding ($g_a$ and $g_n$)  being learned using ‌Binary Cross Entropy and image-level label $y$:
\[
L_{\text{global}} = \text{BCE}\left(y, p_a(g^+_i, G) \right).
\]
For local loss, we use Focal~\cite{focalloss} and Dice~\cite{diceloss} loss at pixel-level. For output feature map $Z \in Z^M$, we compute normal and anomaly maps $S^Z_n$ and $S^Z_a$ based on the equation~\ref{eq:likelihood}, then apply bilinear upsampling $\text{Up}(\cdot)$ to match the anomaly mask $S \in \mathbb{R}^{H \times W}$:
\[
\scalebox{0.92}{$
L^Z_{\text{local}} =
\text{Focal}\left( \text{Up}([S^Z_n, S^Z_a]), [1-S, S] \right)
+ \text{Dice}\left( \text{Up}([S^Z_n, S^Z_a]), [1-S, S] \right).
$}
\]
Finally, we combine both terms using $\lambda$ as a weighting factor controlling the contribution of the local loss.
\[
L_{\text{total}} = L_{\text{global}} + \lambda \sum_{Z\in Z^M} L_{\text{local}}^{Z}.
\]

\vspace{-2em}
\section{Inference} 
For each input image $x$, after computing the local features
$Z \in Z^M$, the anomaly-aware global embedding $g_i^+$, and textual
embeddings $G=\{g_a,g_n\}$, the image-level anomaly score and
low-resolution anomaly maps are calculated as:
\[
\hat{y}=p_a(g_i^+,G),
\qquad
\hat{S}^{Z}_{a,(j,k)}=p_a(Z_{j,k},G),
\qquad Z\in Z^M.
\]
The branch-level anomaly maps are then averaged as:
\[
\hat{S}_a =
\frac{1}{|Z^M|}
\sum_{Z\in Z^M}\hat{S}_a^Z,
\]
followed by bilinear upsampling and Gaussian smoothing to obtain the
final anomaly map $\hat{S}$.

%% file: sec/4-experiments_1.tex
\vspace{-1em}
\section{Experiments}
\subsection{Experiments Setting}
\label{sec:Experiments_Setting}
\paragraph{Datasets.}  
To ensure a comprehensive evaluation, we conduct experiments on 14 real-world anomaly detection datasets spanning industrial defect detection and medical abnormality analysis across 
diverse domains and anomaly types.  
For industrial anomaly detection, we utilize MVTec AD~\cite{bergmann2019mvtec}, VisA~\cite{zou2022spot-visa}, DTD-Synthetic~\cite{aota2023zero-dtd}, SDD~\cite{tabernik2020segmentation-sdd}, 
BTAD~\cite{mishra2021vt-btad}, DAGM~\cite{wieler2007weakly-dagm}, and MPDD~\cite{jezek2021deep-mpdd}, which primarily focus on texture and structural defects in manufactured scenarios.  
For medical anomaly detection, we evaluate on ISIC~\cite{gutman2016skin}, CVC-ColonDB~\cite{tajbakhsh2015automated-colondb}, CVC-ClinicDB~\cite{bernal2015wm-clinicdb}, 
TN3K~\cite{gong2021multi-tn3k}, BrainMRI~\cite{BrainMRI}, HeadCT~\cite{HeadCT}, and BR35H~\cite{Br35h}, covering pathological abnormalities across dermoscopic, colonoscopic, retinal, 
and brain imaging domains. 
Following prior works~\cite{zhou2023anomalyclip,cao2024adaclip,chen2023april}, we train our model on the test split of MVTec-AD and evaluate its generalization to the test splits of all other datasets. For evaluation on MVTec-AD itself, we instead train on VisA. MVTec-AD has no category overlap with VisA or the other evaluated datasets, ensuring that all evaluations remain cross-dataset and category-disjoint. Further details on preprocessing and hyperparameters are provided in Appendix~\ref{sec:implementation_details}.

\paragraph{Evaluation Metrics.}  
Following previous studies~\cite{jeong2023winclip,deng2023anovl}, we employ AUROC, AP, and F1-max to evaluate the model’s ability to distinguish between normal and anomalous 
images. For pixel-level anomaly localization, we use AUROC, AUPRO, and F1-max to assess effectiveness in identifying anomalous regions.
For each dataset, we report the average performance across object categories as the dataset-level result.


\vspace{-0.6em}
\subsection{Main Results}
\input{tables/industrial_medical}

\paragraph{Baselines.} We compare our method against two categories of CLIP-based zero-shot anomaly detection (ZSAD) approaches: training-free methods including AnoVL \cite{deng2023anovl} and 
training-required methods which involve training on an auxiliary dataset before inference, such as VAND \cite{chen2023april}, AnomalyCLIP \cite{zhou2023anomalyclip}, AdaCLIP \cite{cao2024adaclip}, and AA-CLIP \cite{ma2025aa}.
Details on the performance values reported for each model are provided in Appendix \ref{sec:baselines}.

\vspace{-1em}
\paragraph{Zero-shot Performance over Industrial Datasets.} 

We evaluate the generalization of \methodname on seven industrial datasets, as shown in Table~\ref{tab:wo_dino_idustrial}. We report results for both the base model \methodname and enhanced model \methodnamepd ~The industrial image-level results show that both versions outperform prior works by a similar margin. Notably, \methodname improves the state-of-the-art by 2.4\% in AUROC, 4.5\% in AP, and 0.7\% in F1-max on average. These gains highlight the effectiveness of our contributions designed to address the limited sensitivity of global features to local anomalous cues—namely, context-guided prompt learning and anomaly-aware local-to-global fusion.

In the industrial pixel-level results, we observe that \methodname demonstrates consistent and remarkable improvements, achieving 0.2\% higher AUROC, 5.7\% higher AUPRO, and a 1.7\% increase in F1-max on average. \methodnamep pushes the margin even further, improving AUROC by 0.5\%, AUPRO by 9.0\%, and F1-max by 3.8\%, demonstrating that DINOv2 injects complementary knowledge beyond CLIP vision encoder via \dinoattn modules. These improvements are attributed to the architectural modifications designed to preserve spatial alignment—namely, the \clipattn branch in both \methodname and \methodnamepc and the complementary \dinoattn in \methodnamepd

\vspace{-1.1em}
\paragraph{Zero-shot Generalization to Medical Domain.} 
Here, we assess how well the features learned on industrial datasets generalize to domains far from the industrial domain. To this end, we evaluate \methodname on seven medical datasets spanning diverse applications, including skin cancer detection in photography images, colon polyp identification in endoscopy images, thyroid nodule detection in ultrasound images, and brain tumor detection in MRI images. The goal is to determine whether the model has developed a broader understanding of normality and abnormality.

Table~\ref{tab:wo_dino_idustrial} presents the results, showing a consistent image-level trend: both versions outperform prior SOTA by a similar margin, improving average performance by 1.0\% to 3.6\% across reported metrics. At the pixel level on medical datasets, however, the gains are more mixed: \methodname trails AA-CLIP in AUROC and F1-max by $3.5\%$ and $5\%$, respectively, while substantially outperforming it by $13.6\%$ in AUPRO, a more precise metric that is also more commonly used in practice~\cite{bergmann2020uninformed}. With \methodnamepc the gaps in AUROC and F1-max shrink to a subtle margin~($0.6\%$ and $1.6\%$), while it outperforms AA-CLIP even further by $19\%$ in AUPRO, demonstrating strong zero-shot cross-domain generalization and highlighting the practical benefits of the proposed methods over previous methods in real-world applications.

\vspace{-.8em}
\subsection{Ablation Study}
\vspace{-.2em}

To assess the effectiveness of each contribution, we conduct a series of controlled ablation experiments. 
All models are trained for five epochs, with only a single component modified at each stage while keeping all other factors unchanged. 
We report F1-max and AUROC scores on VisA and MVTec-AD for both image- and pixel-level performance.

\input{tables/ablations}

\vspace{-1em}
\paragraph{Anomaly-aware Local-to-Global Fusion.}
A purely global image representation from the CLIP image encoder can be too coarse, as CLIP is trained to align images with relatively coarse text descriptions \cite{salehi2023time, oquab2023dinov2}. In our anomaly detection setting, the distributions of normal and abnormal inputs can be very close, as illustrated in Figures 4, 5, and 8 of the appendix, which leads to highly similar global representations. To address this limitation, we fuse local dense features containing abnormal regions with the global representation, enabling the model to learn more discriminative prompts. Moreover, since the informative cues may be confined to only a few dense patches while large parts of the image are irrelevant or distracting, we first score the patches and then fuse only the most relevant ones. This motivates our combination of global and local information: the global representation preserves the overall semantic context of the image, while the local features allow the model to focus on fine-grained regions where the critical evidence appears. In this way, the two representations play complementary roles. As shown in Table \ref{table:abl_sem_patc_aggr}, aggregating anomalous local embeddings into the global \texttt{[CLS]} token improves F1-max by 1.9\% on average across VisA and MVTec-AD datasets.


\vspace{-1em}
\paragraph{Context-guided Prompt Learning.}
Context-guided prompt learning uses the global visual token only during training, where the normal and abnormal prompts are conditioned on the image context before being optimized. This exposes the learnable textual prototypes to source-domain visual variation, rather than learning them from fixed textual tokens alone. To evaluate its effect, we compare training with and without concatenating $g_i$, the visual [CLS] token, to the input prompts. As shown in Table~\ref{table:abl_img_cnd_pmt}, context-guided prompt learning improves both image-level and pixel-level performance, increasing F1-max by 0.9\% on average across the reported settings.

To understand the mechanism of this improvement, we analyze the geometry of the learned normal and abnormal textual prototypes in Table (a) of Figure~\ref{fig:cg_analysis}; the metric definitions are provided in Appendix~\ref{app:cg_geometry}. Context guidance increases the separation between $g_n$ and $g_a$, indicating that the learned normal and abnormal prototypes become more distinct, despite receiving the same image context during training. It also improves the alignment between the textual abnormality direction, $g_a-g_n$, and the corresponding visual abnormality directions at both image and patch levels. Moreover, the anomaly-margin gap increases, indicating better separation between normal and anomalous samples or patches in the learned text-induced scoring space. The qualitative maps in Figure~\ref{fig:cg_analysis}(b) show a similar trend: with context guidance, responses become less diffuse over normal regions and more concentrated around abnormal evidence. These results suggest that context-guided prompt learning improves the visual grounding of the learned textual prototypes, while dense localization is further refined by the attention branches.

\vspace{-1em}
\paragraph{Extended Self-Correlation Attention.} 
In Table \ref{table:abl_qkv2_attn} the effect of using separate attention tokens ($qq$, $kk$, or $vv$) and their combinations ($qq+kk$ or $qq+kk+vv$) for self-correlation attention is examined. Extending self-correlation to include all three tokens ($qq+kk+vv$) improves pixel-level F1-max by 1.7\% and image-level F1-max by 0.5\% over the $vv$-only baseline. These results suggest that $qq$, $kk$, and $vv$ capture certain spatial information exclusively, therefore combining them improves alignment. 

\vspace{-1em}
\paragraph{DINO-guided Spatial Attention.} %
Table~\ref{table:abl_ens_attn} evaluates \dinoattn by comparing the adapted CLIP branch alone with standalone \dinoattn variants and their combination. The adapted CLIP branch, \clipattn, performs better than standalone \dinoattn for image-level classification, which is expected because it remains directly tied to CLIP's text-aligned representation space. However, combining \dinoattn with \clipattn improves pixel-level F1-max by 1.8\% and image-level F1-max by 0.3\% on average over \clipattn alone. This indicates that \dinoattn contributes complementary spatial information rather than replacing the CLIP-derived branch. 
The qualitative comparison in Fig.~\ref{fig:branch_attention_comparison} further illustrates complementarity roles of the attention branches. we visualize anomaly maps from the VV attention used in AnomalyCLIP, our proposed \clipattn, \dinoattn, and their default aggregation. VV often captures coarse anomalous evidence but remains spatially weak. \clipattn produces more semantically constrained and compact responses, which helps suppress distracting normal regions and improves localization of small defects. However, this semantic selectivity can under-cover spatially extended or visually heterogeneous anomalies. In contrast, \dinoattn emphasizes local visual similarity and spatial grouping, allowing it to recover broader anomaly structure and fine local continuity. Since this signal is less directly constrained by CLIP’s text-aligned semantics, \dinoattn may also activate visually similar normal regions. Aggregating \clipattn and \dinoattn balances these effects: \clipattn provides semantic filtering, while \dinoattn contributes spatial continuity. As a result, the combined map more consistently highlights the true anomalous regions and reduces branch-specific false positives or false negatives. For further evaluation on using SAM~\cite{sam} and DINOv3 instead of DINOv2 in \dinoattn, refer to Appendix \ref{sec:DIVNO_SAM_COMP}.

\begin{figure*}[tphb]
\centering

\begin{tabular}{@{}c@{}}

\begin{minipage}{0.96\textwidth}
\centering
\vspace{-.2em}
\textbf{(a) Text-geometry Analysis}
\vspace{0.3em}

\setlength{\tabcolsep}{3pt}
\small
\resizebox{\linewidth}{!}{%
\vspace{0.4em}
\begin{tabular}{ll c cc cc}
\toprule
\multirow{2}{*}{Dataset} 
& \multirow{2}{*}{Train} 
& \multirow{2}{*}[-0.3em]{\shortstack{Text\\sep.}}
& \multicolumn{2}{c}{Patch-level} 
& \multicolumn{2}{c}{Image-level} \\
\cmidrule(lr){4-5}
\cmidrule(lr){6-7}
& & 
& \shortstack{Abnormal\\direction align.} 
& \shortstack{Margin\\gap} 
& \shortstack{Abnormal\\direction align.} 
& \shortstack{Margin\\gap} \\
\midrule

\multirow{3}{*}{MVTec-AD}
& w/o CG 
& 1.6065 
& 0.2964 
& 0.1891 
& 0.2860 
& 0.1236 \\

& with CG 
& 1.7992 
& 0.3493 
& 0.2388 
& 0.2904 
& 0.1309 \\

& gain 
& +12.0\% 
& +17.8\% 
& +26.3\% 
& +1.5\% 
& +5.9\% \\

\midrule

\multirow{3}{*}{VisA}
& w/o CG 
& 1.8064 
& 0.3193 
& 0.2639 
& 0.1979 
& 0.0476 \\

& with CG 
& 1.8650 
& 0.3655 
& 0.3040 
& 0.2261 
& 0.0535 \\

& gain 
& +3.2\% 
& +14.5\% 
& +15.2\% 
& +14.2\% 
& +12.5\% \\

\bottomrule
\end{tabular}
}
\end{minipage}

\\

\begin{minipage}{0.99\textwidth}
\centering
\vspace{0.8em}
\textbf{(b) Qualitative Analysis}\\
\vspace{0.3em}
\includegraphics[width=\linewidth]{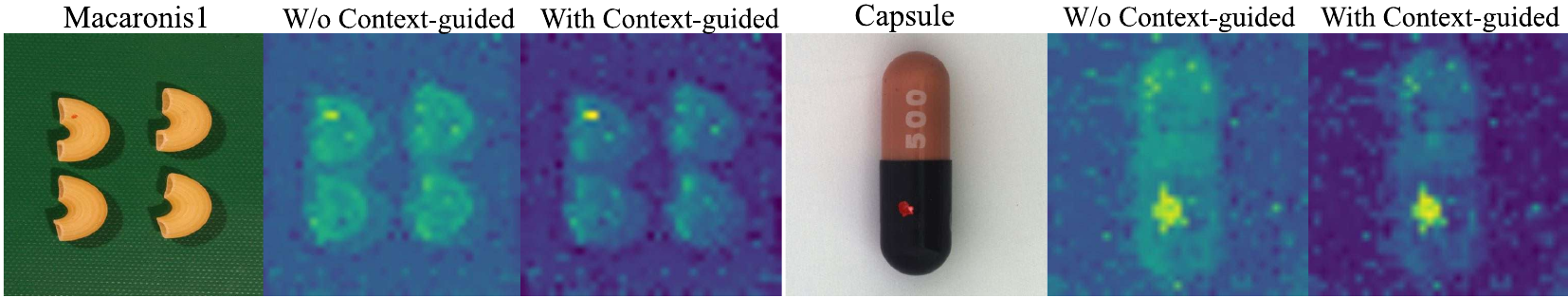}
\end{minipage}

\end{tabular}

\vspace{0.4em}
\caption{\textbf{Effect of context-guided prompt learning.}
(a) Geometry of the learned normal and abnormal textual prototypes. Context guidance increases text separation, abnormal-direction alignment, and anomaly-margin gaps at both patch and image levels. 
(b) Qualitative localization maps with and without context guidance. With context guidance, responses become less diffuse over normal regions and more concentrated around abnormal evidence.}
\label{fig:cg_analysis}
\vspace{-1.7em}
\end{figure*}


\begin{figure}[htbp]    
    \centering
\includegraphics[width=0.95\textwidth]{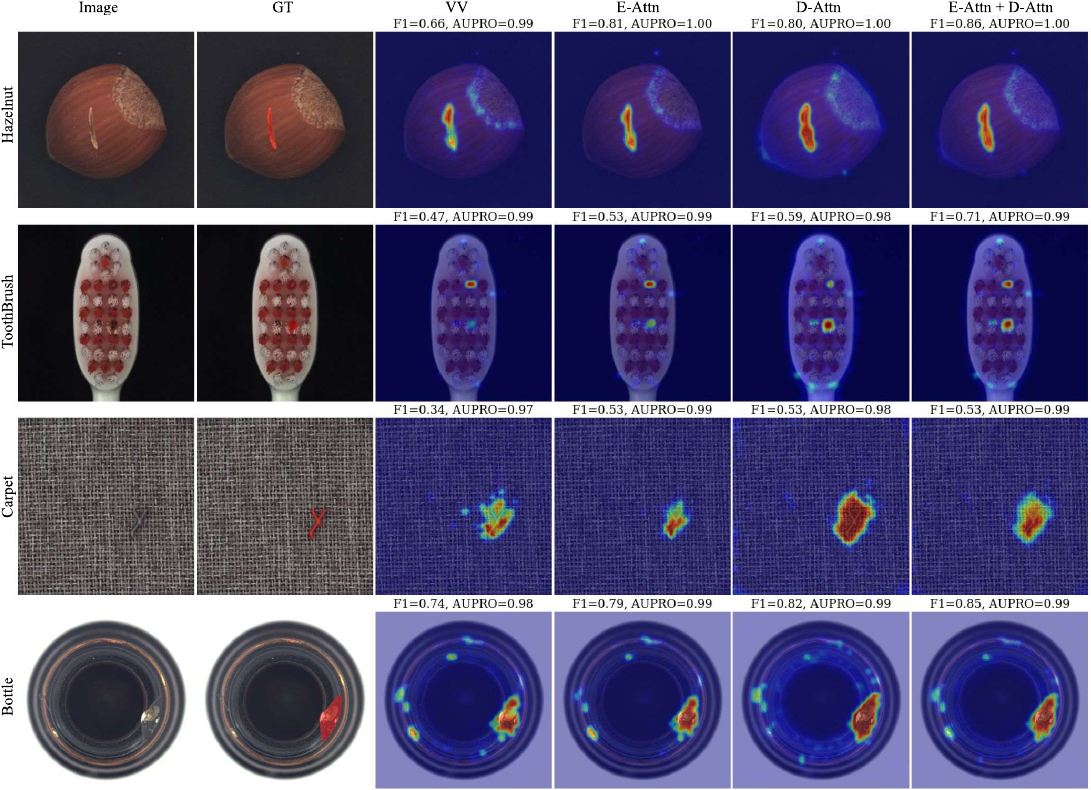}
    \vspace{-0.8em}
    \caption{\textbf{Qualitative comparison of dense attention branches.} VV attention, used in AnomalyCLIP, often gives coarse or noisy responses. \clipattn improves semantic selectivity, while \dinoattn introduces complementary patch-level visual similarity and spatial continuity. Hence, their combination suppresses branch-specific artifacts and yields the best F1 scores.}
    \vspace{-1.5em}
\label{fig:branch_attention_comparison}

\end{figure}
\subsection{Computational Analysis}
\label{Computational Analysis}
\vspace{-.3em}

We evaluate the computational efficiency of \methodname against competing methods in terms of training time, inference throughput, and memory usage (Table~\ref{table:abl_efficiency}) using a single A6000 GPU. AA-CLIP achieves the highest inference FPS; however, it lags substantially behind other models in mean industrial performance. Among the rest, both variants of our method operate with comparable or lower computational cost than others, while achieving large performance gains, indicating a better performance–efficiency trade-off. The modest impact on inference speed in \methodnamep stems from a parallel design in which DINOv2 features are extracted concurrently with CLIP, avoiding additional sequential overhead.

\begin{table}[htbp]
\centering
\caption{\textbf{Computational efficiency comparison.} For each model, we report training time per epoch, inference throughput (FPS), and inference memory on MVTec-AD, along with mean pixel-level AUPRO and image-level AP over industrial datasets.}
\vspace{0.4em}
\label{table:abl_efficiency}
\begin{tabular}{lccccc}
    \toprule
    Model & \makecell{Train\\(min/epc)} & \makecell{Infer.\\(FPS)} & \makecell{Mem.\\(GB)} & \makecell{Mean Perf.\\(P-AUPRO, I-AP)} \\
    \midrule
    AdaCLIP     & 9.36 & 3.43 & 4.21 & 50.1, 89.2  \\
    AA-CLIP     & 7.55 & \textbf{5.12} & 5.19 & 42.0, 85.2 \\
    AnomalyCLIP & \textbf{7.21} & \underline{3.99} & \textbf{3.76} & 83.3, 88.6 \\
    \methodname & \underline{7.38} & 3.67 & \underline{3.81} & \underline{89.0}, \textbf{93.7} \\
    \methodnamep      & 8.15 & 3.38 & 4.20 & \textbf{92.3}, \underline{93.5} \\
    \bottomrule
\end{tabular}%
\vspace{-1.5em}
\end{table}

\subsection{Visual Analysis} \label{sec:Visual_Analysis}
\begin{figure}[H]    
    \centering
     \vspace{-1em}   \includegraphics[width=1\textwidth]{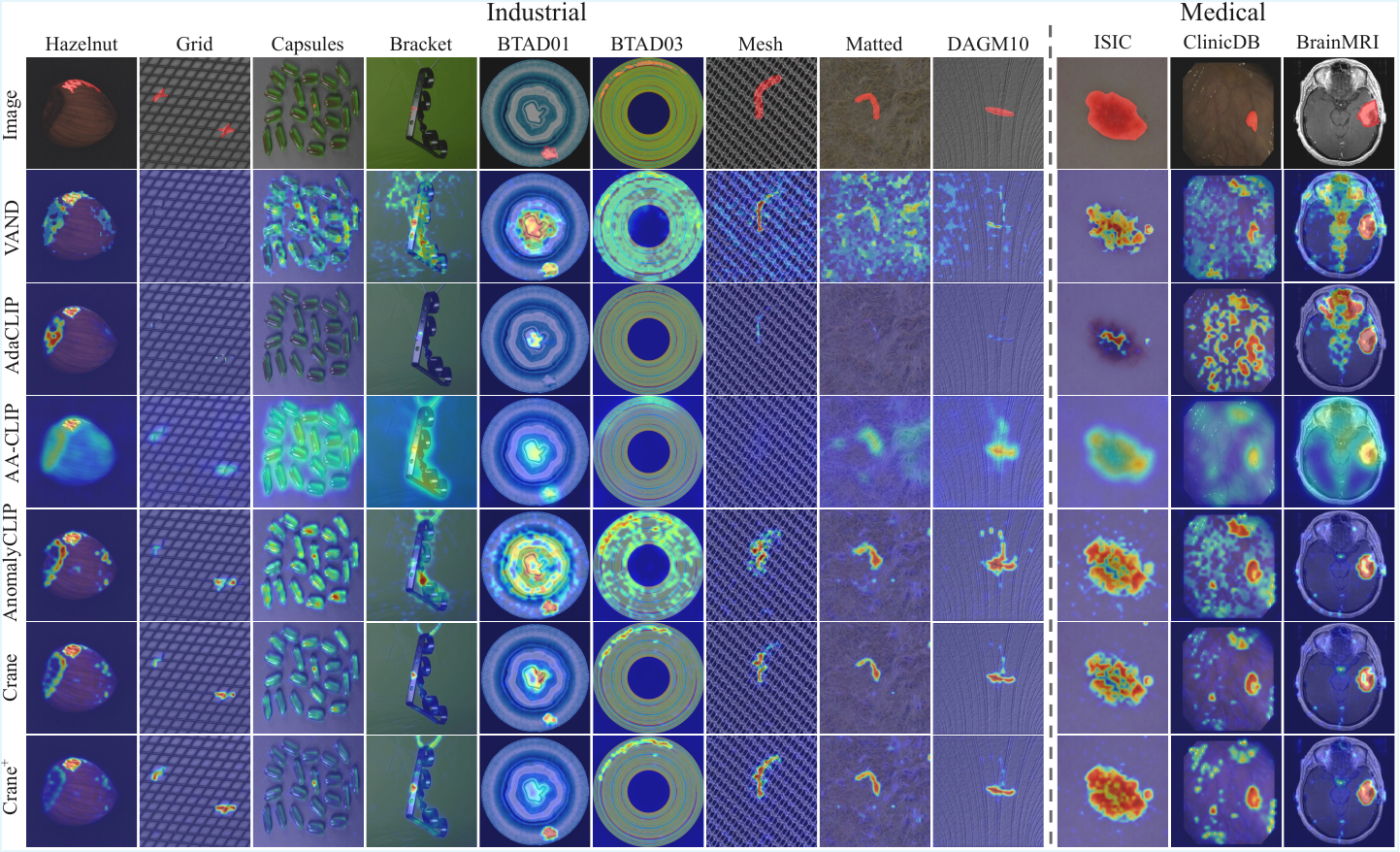}
    \vspace{-1.8em}
    \caption{\textbf{Qualitative localization comparison}. Models are trained on industrial datasets as described in section \ref{sec:Experiments_Setting}. By leveraging stronger semantic correlations among patches, \methodname achieves remarkable reductions in both false positive and false negative rates compared to prior methods, resulting in cleaner and more precise anomaly localization. This advantage is further amplified in \methodnamepc demonstrating state-of-the-art localization performance.}
    \vspace{-1em}
\label{fig:crane_qualitative_comparison}
\end{figure}

\vspace{-0.3em}
To provide an intuitive comparison, we present anomaly maps generated by the top competing models: VAND, AdaCLIP, AA-CLIP, AnomalyCLIP, and both versions of our model, \methodname and \methodnamepc across industrial images from MVTec-AD, VisA, MPDD, BTAD, DAGM, and DTD-Synthetic, as well as medical images from ISIC, CVC-ClinicDB, and BrainMRI. Note that BrainMRI only has image-level labels, and the provided sample is annotated by a medical professional.
As shown in the Figure~\ref{fig:crane_qualitative_comparison},
 VAND suffers from high false positive rates (FPR), while AdaCLIP exhibits a high false negative rate (FNR). AA-CLIP and AnomalyCLIP improve sensitivity by reducing FNR but still struggle with high FPR. By leveraging stronger semantic correlations among patches, \methodname reduces both FPR and FNR over previous methods, yielding cleaner and more precise results. This effect is further enhanced in \methodnamepc demonstrating superior localization performance.

\section{Conclusion} 
\vspace{-0.5em}
In this work, we introduced \methodname for zero-shot anomaly detection. Our approach addresses major challenges of CLIP in this task, namely the global features' limited sensitivity to local anomalous patterns and spatial misalignment. For the first one, we guide prompts using the global visual token and fuse anomaly-relevant local patches into the global token. For the latter, we propose an extended self-correlation attention mechanism, and for more complex domains, DINO-guided spatial attention with relatively low computational overhead. The extensive experiments across 14 medical and industrial datasets demonstrate the effectiveness of the four components in tackling the two limitations, achieving state-of-the-art performance in zero-shot anomaly detection and localization. \mytextcolor{orange}{More discussion on limitations in Appendix~\ref{sec:category_level_qualitative_results}.}

\section*{Acknowledgement}

Mohammad Sabokrou was supported by JSPS KAKENHI Grant-in-Aid for Scientific Research (B), Grant Number JP26K02988, and JSPS KAKENHI Grant-in-Aid for Early-Career Scientists, Grant Number 24K20806.

%% file: tables/industrial_medical.tex

\newcommand{\topone}[1]{\textbf{#1}}
\newcommand{\secondbest}[1]{\underline{#1}}

\definecolor{industrialColor}{RGB}{65, 105, 225}
\definecolor{medicalColor}{RGB}{0, 128, 128} 

\begin{table*}[t!]
\centering
\caption{\textbf{Comparison of ZSAD methods across industrial and medical domains.} We compare our method against the state-of-the-art across 14 diverse industrial and medical datasets. The best performance is highlighted in \textbf{bold}, while the second-best is \underline{underlined}. A $^\dag$ symbol next to a method name indicates training-free models. On \textcolor{industrialColor}{industrial benchmarks}, unlike AnomalyCLIP and AdaCLIP, which fail to achieve consistent improvements, both versions of our model enhance the state-of-the-art in image-level and pixel-level metrics. On \textcolor{medicalColor}{medical benchmarks}, our method achieves competitive performance or significant improvement compared to the SOTA as shown by image and pixel metrics.}
\vspace{.5em}
\label{tab:wo_dino_idustrial}
\resizebox{1\linewidth}{!}{
\begin{tabular}{@{}ccccccccc@{}}        
\toprule[1.5pt]
Metric &
Dataset &
  {AnoVL$^{\dag}$} &
  {VAND} &
  {AnomalyCLIP} &
  {AdaCLIP} & 
  {AA-CLIP} &
  {\textbf{Ours (\methodname)}} &
  {\textbf{Ours \methodnamepp}} \\ \midrule
\multirow{12}{*}{\begin{tabular}[c]{@{}c@{}}Image-level $\uparrow$ \\ \footnotesize{(AUROC, AP, F1-max)}\end{tabular}} &
 \textcolor{industrialColor}{MVTec} &
    (92.5, 95.1, 93.2) & 
    (86.1, 93.5, 88.9) & 
    (91.5, 96.2, 92.7) &
    (89.2, 95.7, 90.6) &  
    (90.5, 95.4, 90.7) &
    (\secondbest{93.8}, \secondbest{97.5}, \topone{93.8}) &
    (\topone{93.9}, \topone{97.6}, \secondbest{93.6}) \\
 & \textcolor{industrialColor}{VisA} &
    (79.2, 80.2, 79.7) &
    (78.0, 81.4, 80.7) &
    (82.1, 85.4, 80.4) &
    (\topone{85.8}, 79.0, \topone{83.1}) & 
    (84.6, 82.9, 78.8) & 
    (\secondbest{85.3}, \topone{87.9}, \secondbest{82.6}) &
    (83.6, \secondbest{86.7}, 81.2) \\ 
 & \textcolor{industrialColor}{MPDD} &
    (72.7, 83.6, \topone{88.3}) & 
    (73.0, 80.2, 76.0) &
    (77.0, 82.0, 80.4) &
    (76.0, 80.2, 82.5) & 
    (74.7, 76.8, 79.7) & 
    (\topone{81.4}, \topone{84.8}, 81.6) &
    (\secondbest{81.0}, \secondbest{84.1}, \secondbest{83.0}) \\
 & \textcolor{industrialColor}{BTAD} &
    (80.3, 72.8, 73.0) & 
    (73.6, 68.6, 82.0) &
    (88.3, 87.3, 83.8) &
    (88.6, 93.8, 88.2) & 
    (\secondbest{94.8}, \topone{97.3}, \topone{93.7}) &
    (94.4, 95.9, \secondbest{91.7}) &
    (\topone{96.3}, \secondbest{97.0}, \topone{93.7}) \\
 & \textcolor{industrialColor}{KSDD} &
    (94.4, 90.8, 88.0) & 
    (79.8, 71.4, 85.2) &
    (84.7, 80.0, 82.7) & 
    (\secondbest{97.1}, 89.6, \secondbest{90.7}) &  
    (86.8, 59.5, 55.9) & 
    (\topone{97.8}, \secondbest{94.3}, \topone{91.6}) &
    (\topone{97.8}, \topone{94.5}, 89.7) \\
 & \textcolor{industrialColor}{DAGM} &
    (89.7, 76.1, 74.7) & 
    (94.4, 83.8, 91.8) &
    (97.5, 92.3, 90.1) &
    (\secondbest{99.1}, 88.5, \topone{97.5}) & 
    (95.0, 85.3, 82.6) & 
    (\topone{99.2}, \topone{97.4}, \secondbest{95.8}) &
    (98.9, \secondbest{96.1}, 94.7) \\ 
 & \textcolor{industrialColor}{DTD} &
    (94.9, 93.3, \topone{97.3}) & 
    (94.6, 95.0, \secondbest{96.8}) &
    (93.5, 97.0, 93.6) &
    (95.5, 97.3, 94.7) & 
    (91.3, 97.0, 92.8) &
    (\topone{96.3}, \topone{98.5}, 95.3) &
    (\secondbest{95.8}, \secondbest{98.2}, 94.6)\\ \cmidrule{2-9}

 & \textcolor{industrialColor}{Average} &
    (86.2, 84.6, 84.9) & 
    (81.6, 82.0, 85.9) &
    (87.8, 88.6, 86.2) & 
    (90.2, 89.2, 89.6) &
    (88.3, 85.2, 81.9) &
    (\topone{92.6}, \topone{93.7}, \topone{90.3}) & 
    (\secondbest{92.5}, \secondbest{93.5}, \secondbest{90.1}) \\   

\cmidrule{2-9}
& \textcolor{medicalColor}{HeadCT} &
   (82.3, 81.2, 79.1) & 
   (89.2, 89.5, 82.1) &
   (93.4, 91.6, \secondbest{90.8}) &
   (91.8, 90.6, 84.1) & 
   (\secondbest{95.2}, {92.0}, {90.1}) &
   (\topone{95.3}, \topone{95.7}, \topone{91.1}) &
   ({94.6}, \secondbest{95.4}, 89.7) \\
 & \textcolor{medicalColor}{BrainMRI} & 
   (84.3, 89.2, 84.8) & 
   (89.6, 91.0, 88.5) &
   (90.3, 92.2, 90.2) &
   (93.5, 95.6, 89.7) & 
   ({93.0}, 92.1, {91.2}) &
   (\secondbest{95.4}, \secondbest{96.1}, \topone{93.9}) &
   (\topone{96.3}, \topone{97.4}, \secondbest{93.5}) \\
 & \textcolor{medicalColor}{Br35H} & 
   (80.0, 80.7, 75.2) & 
   (91.4, 91.9, 84.2) &
   (94.6, 94.7, 89.1) &
   (92.3, 93.2, 85.3) & 
   ({96.0}, {94.0}, {90.9}) &
   (\secondbest{96.3}, \secondbest{96.8}, \topone{91.7}) &
   (\topone{96.4}, \topone{97.2}, \secondbest{91.0}) \\ \cmidrule{2-9}
 & \textcolor{medicalColor}{Average} & 
    (82.2, 83.7, 79.7) & 
    (90.1, 90.8, 84.9) &
    (92.8, 92.8, 90.0) &
    (92.5, 93.1, 86.4) &
    (\secondbest{94.7}, 92.7, {90.7}) &
    (\topone{95.7}, \secondbest{96.2}, \topone{92.2}) &
    (\topone{95.7}, \topone{96.7}, \secondbest{91.4}) \\
    
    \midrule

\multirow{12}{*}{\begin{tabular}[c]{@{}c@{}}Pixel-level $\uparrow$ \\ \footnotesize{(AUROC, AUPRO, F1-max)}\end{tabular}} 

& \textcolor{industrialColor}{MVTec} &
    (90.6, 77.8, 36.5) & 
    (87.6, 44.0, 39.8) &
    (91.1, 81.4, 39.1) &
    (88.7, 37.8, {43.4}) & 
    (\topone{91.9}, 45.9, \topone{47.0}) &
    (\secondbest{91.3}, \secondbest{84.6}, 41.3) & 
    ({91.2}, \topone{88.1}, \secondbest{43.8}) \\
& \textcolor{industrialColor}{VisA} &
    (85.2, 60.5, 14.6) & 
    (94.2, 86.8, \secondbest{32.3}) &
    (\topone{95.5}, 87.0, 28.3) &
    (\topone{95.5}, 72.9, \topone{37.7}) &
    (\topone{95.5}, 25.0, 30.5) &
    (95.1, \secondbest{87.5}, 30.9) & 
    (\secondbest{95.3}, \topone{90.6}, 30.2) \\
& \textcolor{industrialColor}{MPDD} &
    (62.3, 38.3, 15.6) & 
    (94.1, 83.2, 30.6) &
    (96.5, 88.7, 34.2) &
    (96.1, 62.8, 34.9) & 
    ({96.7}, 26.1, 28.9) &
    (\secondbest{97.0}, \secondbest{89.3}, \secondbest{38.2}) & 
    (\topone{97.6}, \topone{93.2}, \topone{42.0}) \\
& \textcolor{industrialColor}{BTAD} &
    (75.2, 40.9, 23.4) & 
    (60.8, 25.0, 38.4) &
    (94.2, 74.8, 49.7) &
    (92.1, 20.3, 51.7) & 
    (\topone{97.0}, 54.3, {55.6}) &
    ({96.6}, \secondbest{81.3}, \secondbest{56.9}) & 
    (\secondbest{96.7}, \topone{86.8}, \topone{61.1}) \\
& \textcolor{industrialColor}{KSDD} &
    (97.1, 82.6, 23.1) &
    (79.8, 65.1, 56.2) &
    (90.6, 67.8, 51.3) &
    (97.7, 33.8, 54.5) &
    (\secondbest{98.1}, 29.6, 37.4) &
    ({97.9}, \secondbest{95.9}, \secondbest{60.2}) & 
    (\topone{99.2}, \topone{97.4}, \topone{62.4}) \\
& \textcolor{industrialColor}{DAGM} &
    (79.7, 56.0, 12.8) & 
    (82.4, 66.2, 57.9) &
    (95.6, 91.0, 58.9) &
    (91.5, 50.6, 57.5) & 
    ({95.8}, 51.5, 53.7) &
    (\topone{96.3}, \secondbest{91.2}, \topone{67.2}) & 
    (\secondbest{96.2}, \topone{93.8}, \secondbest{66.8}) \\
& \textcolor{industrialColor}{DTD} &
    (97.7, 90.5, 46.8) &
    (95.3, 86.9, \topone{72.7}) &
    (97.9, 92.3, 62.2) &
    (97.9, 72.9, 71.6) &
    (96.5, 61.4, 60.5) &
    (\secondbest{98.3}, \secondbest{93.3}, 68.8) & 
    (\topone{98.8}, \topone{96.0}, \secondbest{71.8}) \\ \cmidrule{2-9}
& \textcolor{industrialColor}{Average} &
    (84.0, 63.8, 24.7) &
    (84.9, 65.3, 46.8) &
    (94.5, 83.3, 46.2) &
    (94.2, 50.1, 50.2) &
    ({95.9}, 42.0, 44.8) &
    (\secondbest{96.1}, \secondbest{89.0}, \secondbest{51.9}) &
    (\topone{96.4}, \topone{92.3}, \topone{54.0}) \\ 

    \cmidrule{2-9}
    & \textcolor{medicalColor}{ISIC} &
       (\secondbest{92.6}, {82.2}, \secondbest{76.6}) &
       (89.5, 77.8, 71.5) & 
       (89.7, 78.4, 70.6) &
       (90.3, 54.7, 72.6) & 
       (\topone{93.8}, \topone{85.3}, \topone{78.4}) &
       (88.1, 75.3, 69.8) &
       ({90.6}, \secondbest{83.2}, {73.4}) \\
     & \textcolor{medicalColor}{ColonDB}  &
       (76.2, 44.1, 26.8) &
       (78.4, 64.6, 29.7) & 
       (81.9, 71.3, {37.3}) &
       (82.6, 66.0, 36.1) & 
       (\secondbest{84.0}, 33.4, \secondbest{39.0}) &
       ({82.5}, \secondbest{73.0}, 36.0) &
       (\topone{86.0}, \topone{78.6}, \topone{40.2}) \\
     & \textcolor{medicalColor}{ClinicDB} &  
       (79.7, 51.4, 36.3) &
       (80.5, 60.7, 38.7) & 
       (82.9, 67.8, 42.1) &
       (82.8, 66.4, 40.9) & 
       (\topone{89.9}, 54.0, \topone{53.8}) &
       ({84.0}, \secondbest{69.3}, {42.5}) &
       (\secondbest{88.3}, \topone{74.5}, \secondbest{47.9}) \\
     & \textcolor{medicalColor}{TN3K} &
       (70.2, 34.4, 32.3) &
       (73.6, 37.8, 35.6) & 
       (\topone{81.5}, \secondbest{50.4}, \topone{47.9}) &
       (76.8, 34.0, 40.7) & 
       (\secondbest{80.4}, 39.4, 42.5) &
       (79.4, 48.8, 44.7) &
       (\secondbest{80.4}, \topone{51.7}, \secondbest{45.5}) \\ \cmidrule{2-9}
     & \textcolor{medicalColor}{Average} &
       (79.7, 53.0, 43.0) & 
       (80.5, 60.2, 43.9) &
       ({84.0}, \secondbest{67.0}, {49.5}) &
       (83.1, 55.3, 47.6) &
       (\topone{87.0}, 53.0, \topone{53.4}) &
       (83.5, 66.6, 48.2) &
       (\secondbest{86.4}, \topone{72.0}, \secondbest{51.8}) \\

    \bottomrule[1.5pt]
\end{tabular}
}
\vspace{-1.5em}

\end{table*}

%% file: tables/ablations.tex
\newcommand{\ablIndDataset}{MVAD}
\newcommand{\ablMedDataset}{CLDB}

\newcommand{\cmark}{\ding{51}} 
\newcommand{\xmark}{\ding{55}} 
\begin{table}[htbp]
    \vspace{-0.4em}
    \caption{\textbf{Ablation analysis of key components.} Performance is reported at image-level (I-ROC, I-F1-max) and pixel-level (P-ROC, P-F1-max) on MVTec-AD and VisA. Higher values indicate improved performance. Default configuration is colored.}
    \label{tab:ablation}
    \centering
    \setlength{\tabcolsep}{3pt}

    {\scriptsize
    
    \begin{tabular}{cc}
        \begin{minipage}[t]{0.48\linewidth}
            \centering
            \input{tables/abl_scorebased_pooling}
        \end{minipage}
        &
        \begin{minipage}[t]{0.49\linewidth}
            \centering
            \input{tables/abl_context_guided}
        \end{minipage}
    \end{tabular}
    
    \begin{tabular}{cc}
        \begin{minipage}[t]{0.48\linewidth}
            \centering
            \input{tables/abl_e_attn}
        \end{minipage}
        &
        \begin{minipage}[t]{0.48\linewidth}
            \centering
            \input{tables/abl_d_attn}
        \end{minipage}
    \end{tabular}

    }
    \vspace{-1.5em}
\end{table}

%% file: tables/abl_scorebased_pooling.tex
\subtable[\textbf{Effect of anomaly-aware local-to-global feature fusion.}\label{table:abl_sem_patc_aggr}]{%

\centering
\vspace{0.5em}

\vspace{1em}

    \begin{tabular}{c cc}
        \toprule
        & \multicolumn{2}{c}{Image-level}  \\ 
        \cmidrule(l){2-3}
        Score & \textbf{\ablIndDataset} & \textbf{VisA} \\
        \midrule
        \xmark  & (90.7, 92.3) & (79.2, 78.9) \\
        \rowcolor{blue!10} 
        \cmark  & (\topone{94.7}, \topone{94.3}) & (\topone{82.6}, \topone{80.6}) \\
        \bottomrule
    \end{tabular}

}

%% file: tables/abl_context_guided.tex
\subtable[\textbf{Effect of Context-guided Prompt Learning.}\label{table:abl_img_cnd_pmt}]{%
\centering
\setlength{\tabcolsep}{2pt}
\vspace{0.5em}
\vspace{1em}

    \begin{tabular}{c cc cc}
        \toprule
          & \multicolumn{2}{c}{Pixel-level} &  \multicolumn{2}{c}{Image-level}  \\ 
         \cmidrule(r){2-3}
         \cmidrule(l){4-5}
         Train & \textbf{\ablIndDataset} & \textbf{VisA} & \textbf{\ablIndDataset} & \textbf{VisA} \\
        \midrule
         \xmark & (91.7, 43.5) & (94.8, 28.1)    &   (94.0, 93.3) & (81.9, 80.2) \\
         \rowcolor{blue!10} 
         \cmark & (\topone{92.1}, \topone{44.7}) & (\topone{95.5}, \topone{29.2})    &   (\topone{94.7}, \topone{94.3}) & (\topone{82.6}, \topone{80.6}) \\
        \bottomrule
    \end{tabular}

}

%% file: tables/abl_e_attn.tex
\subtable[\textbf{Effect of different Self-correlations for E-Attn.}\label{table:abl_qkv2_attn}]{%

\centering
\setlength{\tabcolsep}{2pt}
\vspace{0.5em}
\vspace{1em}

\begin{tabular}{c cc cc}
    \toprule
      & \multicolumn{2}{c}{Pixel-level} &  \multicolumn{2}{c}{Image-level}  \\ 
     \cmidrule(r){2-3}
     \cmidrule(l){4-5}
    Self-cors. & \textbf{\ablIndDataset} & \textbf{VisA} & \textbf{\ablIndDataset} & \textbf{VisA} \\
    \midrule
    kk          & (91.6, 43.5) & (94.9, 27.9) & (93.6, 93.7) & (82.2, 80.4)\\
    vv          & (\secondbest{92.1}, 43.8) & (95.0, 26.7) & (93.8, 93.3) & (\secondbest{82.4}, \topone{80.7})\\
    qq          & (91.8, 43.9) & (95.0, \topone{29.4}) & (\secondbest{94.1}, \secondbest{94.0}) & (80.2, 79.4)\\
    qq+kk       & (91.6, \secondbest{44.0}) & (\secondbest{95.2}, 29.3) & (94.0, 93.9) & (82.3, 80.5)\\
    \rowcolor{blue!10} 
    qq+kk+vv    & (\topone{92.1}, \topone{44.7}) & (\topone{95.5}, \secondbest{29.2}) & (\topone{94.7}, \topone{94.3}) & (\topone{82.6}, \secondbest{80.6})\\
    \bottomrule
\end{tabular}
}

%% file: tables/abl_d_attn.tex
\vspace{-.46em}
\subtable[\hspace{-1em}
\parbox{0.95\linewidth}{\centering
\textbf{Effect of D-Attn.}
{\scriptsize Superscripts $^1$, $^2$, and $^3$ denote CLIP-L14, DINOv1-B8, and DINOv2-B14, respectively.}
}
\label{table:abl_ens_attn}]{%

\centering
\setlength{\tabcolsep}{2pt}
    
    \begin{tabular}{c cc cc}
        \toprule
          & \multicolumn{2}{c}{Pixel-level} &  \multicolumn{2}{c}{Image-level}  \\ 
         \cmidrule(r){2-3}
         \cmidrule(l){4-5}
        Branch & \textbf{\ablIndDataset} & \textbf{VisA} & \textbf{\ablIndDataset} & \textbf{VisA} \\
        \midrule
        E-Attn$^{1}$                & (91.2, 40.8) & (\secondbest{94.3}, \topone{29.5}) & (\secondbest{93.7}, \secondbest{93.9}) & (\topone{82.7}, \secondbest{80.4}) \\  
        D-Attn$^{2}$               & (91.5, \secondbest{43.7}) & (94.0, 27.1) & (91.1, 92.9) & (78.4, 78.9) \\
        D-Attn$^{3}$              & (\secondbest{92.0}, 42.7) & (93.7, 26.7) & (91.6, 92.6) & (78.3, 78.3) \\
        \rowcolor{blue!10} 
        Both$^{1+3}$     & (\topone{92.1}, \topone{44.7}) & (\topone{95.5}, \secondbest{29.2}) & (\topone{94.7}, \topone{94.3}) & (\secondbest{82.6}, \topone{80.6}) \\
        \bottomrule
    \end{tabular}
    
}

%% file: sec/5-appendix_1.tex

\clearpage
\appendix
\section*{Appendix}

\renewcommand\thesection{\Alph{section}}
\setcounter{section}{0}

\newlength{\TableCaptionGap}
\setlength{\TableCaptionGap}{0.7em}

\newlength{\FigureCaptionGap}
\setlength{\FigureCaptionGap}{0.2em}

\section{Implementation Details} 
\label{sec:implementation_details}
In this study, we use the publicly available pre-trained CLIP (ViT-L/14@336px)\footnote{\url{https://github.com/mlfoundations/open_clip}} as the default backbone and pre-trained 
DINOv2-B14\footnote{\url{https://github.com/facebookresearch/dinov2}} for further spatial alignment.   
Each input prompt is assigned 12 learnable token embeddings, while 4 learnable deep token embeddings per layer are inserted into the first 9 layers of the text encoder. Moreover, E-attn is applied to the last 20 layers of CLIP's ViT. We use the Adam optimizer with a learning rate of 0.001 and \texttt{betas=(0.6, 0.999)}, and train the model once using a batch size of 8 and the local-loss weighting factor of $\lambda=10$. The random seed for all algorithms is set to 111 to ensure reproducibility. Input images to the vision tower are resized to 518×518 and undergo the same normalization during both training and inference. All experiments are conducted using PyTorch 1.13.1 on a single NVIDIA A6000 GPU.

To evaluate zero-shot anomaly detection performance, we train \methodname on the test set of MVTec-AD and assess its generalization across other datasets.  
For evaluation on MVTec-AD, the model is trained on the test data of VisA. MVTec-AD and VisA contain distinct object categories with no overlap with the samples in other datasets, and the benchmark datasets used in this study span diverse domains, ensuring minimal category overlap across datasets.

\section{Additional Ablations}
\subsection{Incremental Ablations}
Here we show the ablation table in an incremental way, starting from the AnomalyCLIP baseline and sequentially adding each proposed component. As shown, context-guided prompt learning improves both image- and pixel-level performance through better input-dependent alignment; extended self-correlation attention mainly improves pixel-level performance by refining spatial features; score-based aggregation primarily boosts image-level performance by enhancing the global representation while leaving dense predictions unchanged; and \dinoattn provides the final refinement, especially for localization.

\input{tables/ablations_increamental}

\subsection{Simple Anomaly Score Aggregation vs. Score-based Local-to-global Feature Fusion}
   We further compare the introduced score-based feature aggregation with simple score aggregation. To do so, in \methodname, we replace the score-based local-to-global feature fusion with score aggregation, in which instead of mean anomaly-aware feature, mean anomaly score of patches is calculated and averaged with image-level anomaly score. As shown in the table~\ref{tab:score_based_ablation}, aggregating features yields higher performance than simple score aggregation, suggesting that aggregated anomaly score is less effective in capturing anomalous regions that might exist in a picture, as it mainly represents the most frequent regions, especially since the most common patches in a picture are often normal, it will be biased towards normality, reducing precision respectively. Whereas, in feature aggregation the anomalous directions are retained and reflected in the final cosine similarity calculation with the learned anomalous features. Moreover, local-to-global feature aggregation not only amplifies anomalous cues that are already captured by the image-level feature, but also incorporates anomalous regions ignored by the less anomaly sensitive attention of the original CLIP, providing richer and more holistic image-level features.

\begin{table*}[t!]

\centering

\caption{\textbf{Effect of anomaly-aware local-to-global feature fusion.} The local-to-global feature aggregation yields higher performance suggesting more effective capture of anomalous cues.} 

{
\vspace{1em}
   \label{tab:score_based_ablation}

   \begin{tabular}{c cc}

       \toprule

       & \multicolumn{2}{c}{Image-level}  \\

       \cmidrule(l){2-3}

       Approach & \textbf{MVTec-AD} & \textbf{VisA} \\

       \midrule

       Simple Score Aggregation  & (92.7, 96.4, 93.2) & (84.1, 87.2, 81.8) \\

       \rowcolor{blue!10} 

       Score-based Feature Aggregation  & (\topone{93.8}, \topone{97.5}, \topone{93.8}) & (\topone{85.3}, \topone{87.9}, \topone{82.6}) \\

       \bottomrule

   \end{tabular}

}
\end{table*}

\subsection{Details of Geometric Analysis for Context-Guided Prompt Learning}
\label{app:cg_geometry}

We conduct this analysis to examine how context-guided prompt learning affects the learned normal and abnormal textual prototypes. The analysis is performed after training and uses image-level labels and pixel-level masks only for evaluation; no additional supervision is introduced during model optimization or inference. We compare two models trained under the same setting, with and without context-guided prompt learning, and analyze the geometry of their learned textual prototypes.

Let $g_n$ and $g_a$ denote the learned normal and abnormal textual prototypes, respectively. All visual and textual features are $\ell_2$-normalized before cosine-based measurements. We first quantify the separation between the two textual prototypes as
\begin{equation}
\mathrm{Sep}
=
1-\cos(g_n,g_a).
\end{equation}
A larger value indicates that the learned normal and abnormal prototypes are more distinct.

We next define the textual abnormality direction as
\begin{equation}
d_t
=
\frac{g_a-g_n}{\|g_a-g_n\|_2},
\label{eq:text_abnormal_direction}
\end{equation}
which is well-defined when $g_a \neq g_n$. This direction characterizes how the learned textual representation moves from the normal prototype toward the abnormal prototype.

For the image-level analysis, let $v_i$ denote the global visual feature of image $i$. For each category $c$, we compute the normal and anomalous visual prototypes as
\begin{equation}
\mu_c^N
=
\frac{1}{|\mathcal{D}_c^N|}
\sum_{i\in \mathcal{D}_c^N} v_i,
\qquad
\mu_c^A
=
\frac{1}{|\mathcal{D}_c^A|}
\sum_{i\in \mathcal{D}_c^A} v_i,
\end{equation}
where $\mathcal{D}_c^N$ and $\mathcal{D}_c^A$ denote the sets of normal and anomalous images for category $c$, respectively. The image-level visual abnormality direction is then defined as
\begin{equation}
d_c^{\mathrm{img}}
=
\frac{\mu_c^A-\mu_c^N}
{\|\mu_c^A-\mu_c^N\|_2},
\label{eq:image_abnormal_direction}
\end{equation}
which points from the normal visual prototype to the anomalous visual prototype for category $c$.

We measure the alignment between textual and image-level visual abnormality directions by
\begin{equation}
\mathrm{Align}_{\mathrm{img}}
=
\frac{1}{|\mathcal{C}|}
\sum_{c\in\mathcal{C}}
\cos\left(d_t,d_c^{\mathrm{img}}\right),
\label{eq:image_alignment}
\end{equation}
where $\mathcal{C}$ denotes the set of evaluated categories. This category-wise averaging provides a category-balanced estimate of the alignment. A larger value indicates that the textual abnormality direction is more consistent with the visual transition from normal to anomalous images.

For the patch-level analysis, let $z_{i,p}$ denote the patch feature of image $i$ at patch location $p$, and let $S_{i,p}\in\{0,1\}$ denote the corresponding patch-level anomaly label obtained from the ground-truth mask, where $S_{i,p}=1$ indicates an anomalous patch and $S_{i,p}=0$ indicates a normal patch. For each category $c$, we compute the normal and anomalous patch prototypes as
\begin{equation}
\mu_c^{N,\mathrm{patch}}
=
\frac{1}{|\mathcal{P}_c^N|}
\sum_{z\in\mathcal{P}_c^N} z,
\qquad
\mu_c^{A,\mathrm{patch}}
=
\frac{1}{|\mathcal{P}_c^A|}
\sum_{z\in\mathcal{P}_c^A} z,
\end{equation}
where $\mathcal{P}_c^N$ and $\mathcal{P}_c^A$ denote the mask-negative and mask-positive patch sets for category $c$, respectively. The patch-level visual abnormality direction is defined as
\begin{equation}
d_c^{\mathrm{patch}}
=
\frac{
\mu_c^{A,\mathrm{patch}}-\mu_c^{N,\mathrm{patch}}
}
{
\|\mu_c^{A,\mathrm{patch}}-\mu_c^{N,\mathrm{patch}}\|_2
},
\label{eq:patch_abnormal_direction}
\end{equation}
which points from normal patches to anomalous patches in the visual feature space.

The corresponding patch-level abnormal-direction alignment is
\begin{equation}
\mathrm{Align}_{\mathrm{patch}}
=
\frac{1}{|\mathcal{C}|}
\sum_{c\in\mathcal{C}}
\cos\left(d_t,d_c^{\mathrm{patch}}\right).
\label{eq:patch_alignment}
\end{equation}
A larger value indicates stronger agreement between the learned textual abnormality direction and the patch-level visual abnormality direction.

Finally, we evaluate how well the learned textual prototypes separate normal and anomalous visual features using an anomaly margin. For either an image feature $v_i$ or a patch feature $z_{i,p}$, the margin is defined as
\begin{equation}
m(e)
=
\cos(e,g_a)-\cos(e,g_n),
\qquad
e\in\{v_i,z_{i,p}\}.
\label{eq:anomaly_margin}
\end{equation}
A larger margin indicates that feature $e$ is closer to the abnormal textual prototype than to the normal textual prototype.

At the image level, the margin gap is computed as
\begin{equation}
\mathrm{Gap}_{\mathrm{img}}
=
\mathbb{E}_{y_i=1}\left[m(v_i)\right]
-
\mathbb{E}_{y_i=0}\left[m(v_i)\right],
\label{eq:image_margin_gap}
\end{equation}
where $y_i=1$ and $y_i=0$ denote anomalous and normal images, respectively. At the patch level, the margin gap is defined as
\begin{equation}
\mathrm{Gap}_{\mathrm{patch}}
=
\mathbb{E}_{S_{i,p}=1}\left[m(z_{i,p})\right]
-
\mathbb{E}_{S_{i,p}=0}\left[m(z_{i,p})\right].
\label{eq:patch_margin_gap}
\end{equation}
A larger margin gap indicates that anomalous images or patches receive higher abnormal-versus-normal margins than normal ones. Overall, larger text separation, stronger abnormal-direction alignment, and larger margin gaps indicate better visual grounding of the learned normal and abnormal textual prototypes.

\subsection{Additional Ablations on Backbone in \dinoattn}
\label{sec:DIVNO_SAM_COMP}

To further compare the effectiveness of different \dinoattn backbones in \methodnamep, we replace DINOv2-B/14 with SAM-ViT-B and DINOv3-B/16. Table~\ref{table:abl_dino_sam} reports mean performance across the seven industrial datasets. Both DINO variants outperform SAM across the reported pixel- and image-level metrics, indicating that their features are better suited to \dinoattn in this setting. DINOv2 and DINOv3 achieve comparable image-level performance, while DINOv2 performs better at the pixel level.

SAM has lower throughput, partly because its standard image encoder operates at \(1024\times1024\) resolution. DINOv2-B/14 and DINOv3-B/16 both contain 86M parameters, yet DINOv3 has slightly lower throughput in our implementation. For a spatially matched comparison, we use \(592\times592\) inputs for DINOv3-B/16, yielding the same \(37\times37\) patch grid as DINOv2-B/14 at \(518\times518\).

\begin{table}[H]
\centering
\small  
\setlength{\tabcolsep}{3pt}  
\caption{\textbf{Comparison of extra backbones in \dinoattn.} DINOv2 achieves stronger localization and detection averaged across seven industrial datasets, with higher inference throughput.}
\vspace{\TableCaptionGap}
\label{table:abl_dino_sam}
\begin{tabular}{lccc}
    \toprule
    Model & \makecell{Pixel-level\\(ROC, PRO, F1-M)} & \makecell{Image-level\\(ROC, AP, F1-M)} & \makecell{Infr.\\(FPS)} \\
    \midrule
    SAM-ViT-B & 95.7, 89.4, 52.3 & 91.7, 91.7, \secondbest{89.9} & 2.65 \\  
    DINOv3-B16    & \secondbest{96.3}, \secondbest{90.6}, \secondbest{52.8} & \secondbest{92.3}, \secondbest{93.0}, 89.7 & \secondbest{3.29} \\
    DINOv2-B14    & \topone{96.4}, \topone{92.3}, \topone{54.0} & \topone{92.5}, \topone{93.5}, \topone{90.1} & \topone{3.38} \\
    \bottomrule
\end{tabular}
\end{table}

\begin{table*}[!thbp]
\centering
\caption{\textbf{Comparisons of ZSAD methods in the medical domain, Supervised models are trained on medical datasets.} The best performance is \topone{bold}, and the second-best is \secondbest{underlined}.
}
\vspace{\TableCaptionGap}
\label{tab:medical_colondb}
\resizebox{1\linewidth}{!}{
\begin{tabular}{@{}cccccccc@{}}
\toprule[1.5pt]
Metric &
Dataset &
   {WinCLIP$^{\dag}$ } &
   {AnoVL$^{\dag}$ } &
   {VAND} &
   {AnomalyCLIP} &
   {AdaCLIP} &
   {\textbf{\methodnamep}} \\ \midrule  
  \multirow{4}{*}{\begin{tabular}[c]{@{}c@{}}Image-level $\uparrow$\\ \footnotesize{(AUROC, AP, F1-max)}\end{tabular}} 

& HeadCT &
   (81.8, 80.2, 79.8) &
   (82.3, 81.2, 79.1) &
   (89.2, 89.5, 82.1) &
   (\secondbest{93.5}, \secondbest{95.1}, \secondbest{91.7}) &
   (81.5, 85.9 ,75.3) &
   (\topone{96.8}, \topone{97.5}, \topone{92.0}) \\
 & BrainMRI & 
   (86.6, 91.5, 86.3) &
   (84.3, 89.2, 84.8) &
   (89.6, 91.0, 88.5) &
   (\secondbest{95.5}, \secondbest{97.2}, \secondbest{93.5}) &
   (61.5, 73.5 ,76.6) &
   (\topone{97.8}, \topone{98.7}, \topone{95.3}) \\
 & Br35H & 
   (80.5, 82.2, 74.4) &
   (80.0, 80.7, 75.2) &
   (91.4, 91.9, 84.2) &
   (\topone{97.9}, \secondbest{98.0}, \secondbest{92.5}) &
   (52.4, 58.3 ,67.5) &
   (\secondbest{97.3}, \topone{98.1}, \topone{93.7}) \\ \cmidrule{2-8}

 & Average & 
   (83.0, 84.6, 80.2) &   
   (82.2, 83.7, 79.7) & 
   (90.1, 90.8, 84.9) &
   (\secondbest{95.6}, \secondbest{96.8}, \secondbest{92.6}) &
   (65.1, 72.6, 73.1) &
   (\topone{97.3}, \topone{98.1}, \topone{93.7}) \\

    \midrule

\multirow{5}{*}{\begin{tabular}[c]{@{}c@{}}Pixel-level $\uparrow$\\ \footnotesize{(AUROC, AUPRO, F1-max)}\end{tabular}} 

 & ISIC &
   (83.3, 55.1, 48.5) &
   (\topone{92.6}, \secondbest{82.2}, \secondbest{76.6}) &
   (83.1, 70.5, 63.7) &
   (83.0, 63.8, 66.1) &
   (72.8, 3.23, 55.0) &
   (\secondbest{91.5}, \topone{84.1}, \topone{77.6}) \\

 & ColonDB  &
   (70.3, 32.5, 19.6) &
   (76.2, 44.1, 26.8) &
   (\secondbest{88.7}, \secondbest{82.5}, \secondbest{58.8}) &
   (87.5, 78.5, 52.1) &
   (85.5, 24.1, 56.3) &
   (\topone{94.2}, \topone{85.3}, \topone{59.3}) \\

 & ClinicDB &  
   (51.2, 13.8, 24.4) &
   (79.7, 51.4, 36.3) &
   (\topone{93.5}, \topone{86.6}, \topone{71.9}) &
   (\secondbest{92.4}, 82.9, 60.0) &
   (92.3, 54.1, \secondbest{69.6}) &
   (91.3, \secondbest{84.6}, 69.1) \\
 & TN3K &
   (70.7, 39.8, 30.0) &
   (70.2, 34.4, 32.3) &
   (76.9, 37.2, 40.5) &
   (\secondbest{79.2}, \secondbest{47.0}, \secondbest{47.6}) &
   (52.4, 0.45, 23.0) &
   (\topone{87.2}, \topone{54.8}, \topone{49.6}) \\ \cmidrule{2-8}

 & Average &
   (68.9, 35.3, 30.6) &   
   (79.7, 53.0, 43.0) & 
   (\secondbest{85.6}, \secondbest{69.2}, \secondbest{58.7}) &
   (85.5, 68.1, 56.4) &
   (75.8, 20.5, 51.0) &
   (\topone{91.0}, \topone{77.2}, \topone{63.9}) \\

    \bottomrule[1.5pt]
\end{tabular}
}
\end{table*}

\subsection{Ablation Studies on Medical Training}
\label{sec:ablation_medical_training}
Building on the remarkable performance of the model when applied to medical datasets—despite being trained solely on an industrial dataset with significant texture differences—we examine the model's behavior when medical samples are included in training, while the test domain remains unexposed. Following the approach in \cite{zhou2023anomalyclip}, we combine the test split of the classification EndoTect~\cite{endocls} with CVC-ColonDB~\cite{tajbakhsh2015automated-colondb} and evaluate on other medical datasets. For evaluation on CVC-ColonDB, we train the model on CVC-ClinicDB~\cite{bernal2015wm-clinicdb} and EndoTect. Although both CVC-ColonDB and CVC-ClinicDB consist of colonoscopy images, they were captured using different endoscopic equipment, resulting in variations in image quality and texture.

Table~\ref{tab:medical_colondb} presents the performance of the selected models under the aforementioned training scheme. Compared to the default evaluation in Main Table~\ref{tab:wo_dino_idustrial}, Our full model, \methodnamep achieves notable gains in pixel-level anomaly detection, with improvements of 12.1\% in F1-max and 4.6\% in AUROC. At the image level, it records increases of 2.3\% in F1-max and 1.6\% in AUROC. Furthermore, in the current table, \methodnamep surpasses the second-best model by 5.4\% in AUROC and 5.2\% in F1-max for pixel-level performance, and by 1.7\% in AUROC and 1.1\% in F1-max at the image level.

\section{State-of-the-Art Methods}
\label{sec:baselines}  

We compare our proposed model with five state-of-the-art methods from the literature. The reported results are either sourced from the respective papers or reproduced if unavailable. An overview of their approaches and reproduction details is as follows:
\begin{itemize}    
    \item \textbf{AnoVL \cite{deng2023anovl}}: Adapts vision-language models for zero-shot anomaly detection by optimizing model parameters through test-time adaptation. It uses v-v attention~\cite{clipsurgery} to address spatial misalignment of textual and patch embeddings. AUROC, PRO, and F1-Max values on MVTec-AD and VisA are from the original paper; other metrics and datasets are derived from the official code implementation.
    
    \item \textbf{VAND \cite{chen2023april}}: Utilizes a vision encoder fine-tuning strategy with linear projections atop features from an auxiliary training set. Its textual prompting approach is similar to WinCLIP and AnoVL. Metrics for MVTec-AD and VisA are from the original publication. For datasets without official measurements, results are reproduced using default parameters from their paper.
    
    \item \textbf{AnomalyCLIP \cite{zhou2023anomalyclip}}: Introduces object-agnostic learnable prompts for zero-shot anomaly detection. By using a general [object] token in text prompts, it emphasizes anomalous regions across domains, enhancing generalization without category-specific training. The original paper provides metrics for all datasets except F1-Max, which are reproduced using the official codebase and default parameters.
    
    \item \textbf{AdaCLIP \cite{cao2024adaclip}}: Adapts CLIP for zero-shot anomaly detection by incorporating learnable deep tokens into vision and text encoders. It uses static and dynamic prompts; static prompts are shared across images for preliminary adaptation, while dynamic prompts are image-specific. The original paper provides metrics for all datasets except AUPRO and Image-AP, which are reproduced using the official codebase and default parameters.

    \item \textbf{AA-CLIP}~\cite{ma2025aa}: Applies a two-stage adaptation strategy, which first fine-tunes the text encoder via residual adapters while freezing the vision encoder, and then fine-tunes the vision encoder using residual adapters and projectors while freezing the text tower. For datasets (KSDD, DTD-Synthetic, DAGM) and metrics (AUPRO and image-/pixel-level F1-max) that lack official numbers, we reproduce the results using the authors’ default settings and official repository.
\end{itemize}

\section{Datasets}
\newlength{\colwidth}
\settowidth{\colwidth}{normal \& anomalous}
\begin{table*}[!htbp]
    \centering
    \caption{\textbf{Numerical details of the utilized datasets.} $|\mathcal{C}|$ indicates the number of object categories in each dataset. The "Labels" column specifies whether the dataset contains image-level and/or pixel-level annotations.}
    \vspace{\TableCaptionGap}
    \label{tab:datasets_details}
    \resizebox{0.9\linewidth}{!}{%
    \begin{tabular}{lcccccccc}
    \toprule
    \multirow{2}{*}{Category} & \multirow{2}{*}{Dataset} & \multirow{2}{*}{Type} & \multirow{2}{*}{Modalities} & \multirow{2}{*}{$|\mathcal{C}|$} & \multirow{2}{*}{\# Normal \& Anomalous} & \multirow{2}{*}{Detection} & \multicolumn{2}{c}{Task} \\  
    \cmidrule(lr){8-9}
    & & & & & & & detection & localization \\
    \midrule
    \multirow{7}{*}{Industrial} 
    & MVTec AD & Obj \& texture & Photography & 15 & (467, 1258) & Industrial defect & \cmark & \cmark \\ 
    \cmidrule(lr){2-9}
    & VisA & \multirow{4}{*}{Obj} & Photography & 12 & (962, 1200) & Industrial defect  & \cmark & \cmark \\  
    & MPDD & & Photography & 6 & (176, 282) & Industrial defect  & \cmark & \cmark \\  
    & BTAD & & Photography & 3 & (451, 290) & Industrial defect  & \cmark & \cmark \\  
    & SDD & & Photography & 1 & (181, 74) & Industrial defect  & \cmark & \cmark \\  
    \cmidrule(lr){2-9}
    & DAGM & \multirow{2}{*}{Texture} & Photography & 10 & (6996, 1054) & Industrial defect  & \cmark & \cmark \\  
    & DTD-Synthetic & & Photography & 12 & (357, 947) & Industrial defect  & \cmark & \cmark \\  
    \midrule
    \multirow{10}{*}{Medical}  
    & ISIC & Skin & Photography & 1 & (0, 379) & Skin cancer  & \xmark & \cmark \\  
    \cmidrule(lr){2-9}
    & CVC-ClinicDB & \multirow{4}{*}{Colon} & Endoscopy & 1 & (0, 612) & Colon polyp  & \xmark & \cmark \\  
    & CVC-ColonDB & & Endoscopy & 1 & (0, 380) & Colon polyp  & \xmark & \cmark \\  
    & Endo & & Endoscopy & 1 & (0, 200) & Colon polyp  & \cmark & \xmark \\  
    \cmidrule(lr){2-9}
    & TN3K & Thyroid & Radiology (Ultrasound) & 1 & (0, 614) & Thyroid nodule  & \xmark & \cmark \\  
    \cmidrule(lr){2-9}
    & HeadCT & \multirow{3}{*}{Brain} & Radiology (CT) & 1 & (100, 100) & Brain tumor  & \cmark & \xmark \\  
    & BrainMRI & & Radiology (MRI) & 1 & (98, 155) & Brain tumor  & \cmark & \xmark \\  
    & Br35H & & Radiology (MRI) & 1 & (1500, 1500) & Brain tumor  & \cmark & \xmark \\  
    \bottomrule
    \end{tabular}%
    }
\end{table*}
For a comprehensive evaluation of the zero-shot generalization capabilities of the proposed model, we conduct experiments across test set of 14 diverse datasets, 
encompassing two domains (industrial and medical) and three modalities (photography, radiology, and endoscopy).
Table~\ref{tab:datasets_details} provides a detailed listing of the statistical details, application, modality, and anomaly pattern types.As shown, some datasets are solely applicable for anomaly localization or detection because they either contain only abnormal images with segmentation masks (e.g., CVC-ColonDB) or include both normal and abnormal samples but lack pixel-level masks.

\clearpage


\clearpage
\section{Category-level Results}\label{sec:Category_level_Results}
\label{app:category_level_results}
In this section, we report quantitative results for both \methodname and \methodnamep across individual sub-dataset categories, alongside the main aggregated results. This breakdown captures the performance distribution across categories, revealing category-specific behaviors and offering a more nuanced view of model robustness. We also include qualitative results for the full model, \methodnamepc to aid interpretation of outlier cases. All results are reproduced according to the implementation details in Appendix~\ref{sec:implementation_details}.

\subsection{Category-level Quantitative Results} 
\begin{table*}[!thbp] 
    \centering
    \small
    \caption{\textbf{Category-level anomaly localization performance for the dataset BTAD.} each triplet reports (AUROC, AUPRO, F1-max).}
    \vspace{\TableCaptionGap}
    \resizebox{1\linewidth}{!}{
    \begin{tabular}{lccccccc}
    \toprule
    Product & AnVoL & VAND & AnomalyCLIP & AdaCLIP & \textbf{\methodname} & \textbf{\methodnamep} \\ \midrule
    01 & (93.8, 56.0, 46.2) & (89.9, 72.3, 53.3) & (93.7, 73.0, 52.9) & (88.8, 1.60, 54.1) & (96.2, 77.3, 60.8) & (97.0, 84.0, 65.5) \\
    02 & (57.8, 15.8, 17.1) & (86.3, 50.3, 56.7) & (94.4, 66.0, 60.0) & (95.9, 12.0, 64.0) & (95.7, 72.8, 64.0) & (95.0, 81.1, 66.7) \\
    03 & (74.0, 50.9, 6.90) & (91.8, 83.6, 12.0) & (94.6, 87.1, 36.4) & (96.3, 47.0, 38.4) & (97.9, 93.6, 45.9) & (97.9, 95.5, 51.3) \\ \midrule
    \textbf{Mean} & (75.2, 40.9, 23.4) & (89.3, 68.7, 40.6) & (94.2, 75.4, 49.7) & (92.9, 22.8, 48.1) & (96.6, 81.3, 56.9) & (96.7, 86.9, 61.2) \\
    \bottomrule
    \end{tabular}
    }
\end{table*}
\begin{table*}[!htbp] 
    \centering
    \small
    \caption{\textbf{Category-level anomaly classification performance for the dataset BTAD.} each triplet reports (AUROC, AP, F1-max).}
    \vspace{\TableCaptionGap}
    \resizebox{1\linewidth}{!}{
    \begin{tabular}{lccccccc}
    \toprule
    Product & AnVoL & VAND & AnomalyCLIP & AdaCLIP & \textbf{\methodname} & \textbf{\methodnamep} \\ \midrule
    01 & (94.8, 97.9, 92.8) & (82.0, 92.3, 84.3) & (90.9, 96.6, 89.4) & (93.2, 95.6, 91.8) & (98.3, 99.3, 95.7) & (98.2, 99.3, 96.8) \\
    02 & (65.4, 93.8, 93.0) & (82.0, 96.8, 93.5) & (84.1, 97.4, 93.3) & (80.3, 94.8, 87.2) & (86.6, 97.9, 93.7) & (92.2, 98.8, 94.7) \\
    03 & (80.7, 27.6, 33.2) & (57.5, 19.8, 26.4) & (89.8, 70.7, 68.6) & (97.6, 93.6, 85.6) & (98.3, 90.5, 85.7) & (98.5, 92.6, 87.2) \\ \midrule
    \textbf{Mean} & (80.3, 72.8, 73.0) & (73.8, 69.6, 68.1) & (88.2, 87.3, 83.8) & (88.7, 93.8, 88.2) & (94.4, 95.9, 91.7) & (96.3, 96.9, 92.9) \\ 
    \bottomrule
    \end{tabular}
    }
\end{table*}
\clearpage
\begin{table*}[!bp]
    \centering
    \caption{\textbf{Category-level anomaly localization performance for the dataset DAGM.} each triplet reports (AUROC, AUPRO, F1-max).}
    \vspace{\TableCaptionGap}
    \small
    \resizebox{1\linewidth}{!}{
    \begin{tabular}{lccccccc}
        \toprule
        Product & AnVoL & VAND & AnomalyCLIP & AdaCLIP & \textbf{\methodname} & \textbf{\methodnamep} \\ \midrule
        \midrule
        Class01  & (58.0, 18.1, 2.0 )  & (70.5, 52.0, 32.5)  & (88.0, 76.7, 50.2)  & (83.5, 58.8, 44.8)  & (88.3, 72.7, 48.5)  & (86.6, 75.7, 40.2)  \\
        Class02  & (90.0, 75.5, 16.4)  & (88.1, 83.5, 51.1)  & (99.5, 99.1, 64  )  & (96.2, 66.5, 66.6)  & (99.6, 98.2, 73.1)  & (98.7, 98.8, 72.5)  \\
        Class03  & (86.8, 62.2, 7.2 )  & (78.5, 61.5, 36.7)  & (95.9, 93.8, 69.7)  & (96.2, 44.3, 72.7)  & (95.8, 89.9, 73.2)  & (95.6, 93.6, 70.5)  \\
        Class04  & (79.4, 54.7, 8.1 )  & (75.5, 44.1, 5.3 )  & (89.1, 75.3, 35.6)  & (86.7, 17.1, 9.6 )  & (92.5, 80.8, 35.6)  & (93.9, 86.4, 33.6)  \\
        Class05  & (81.8, 56.5, 13.5)  & (82.3, 64.0, 54.9)  & (99.1, 96.9, 74.2)  & (97.5, 52.0, 76.7)  & (99.1, 94.6, 79.5)  & (99.3, 98.1, 80.5)  \\
        Class06  & (93.1, 81.2, 47.6)  & (91.9, 81.7, 74.5)  & (99.1, 96.0, 76.4)  & (99.1, 61.5, 82.7)  & (99.3, 93.2, 78.8)  & (99.8, 99.1, 82.8)  \\
        Class07  & (63.3, 26.9, 3.5 )  & (83.6, 69.6, 54.2)  & (90.3, 86.5, 70  )  & (93.3, 57.4, 72.1)  & (90.5, 87.8, 72.8)  & (89.9, 88.7, 70.9)  \\
        Class08  & (58.6, 23.0, 0.3 )  & (80.2, 64.1, 12.1)  & (98.3, 96.3, 55.7)  & (93.5, 40.1, 64.4)  & (99.0, 98.4, 68.4)  & (99.6, 99.6, 68.7)  \\
        Class09  & (89.2, 70.1, 3.9 )  & (90.6, 78.6, 26.9)  & (98.3, 92.7, 33.5)  & (94.9, 60.8, 44.0)  & (99.9, 99.2, 76.0)  & (99.9, 99.6, 75.2)  \\
        Class10  & (97.0, 91.6, 25.7)  & (83.0, 61.1, 25.5)  & (98.5, 97.1, 60.1)  & (96.1, 47.5, 61.2)  & (98.8, 97.1, 66.5)  & (99.2, 98.9, 73.4)  \\ 
        \midrule
        \textbf{Mean}  & (79.7, 56.0, 12.8)  & (82.4, 66.0, 37.4)  & (95.6, 91.0, 58.9)  & (93.7, 50.6, 59.5)  & (96.3, 91.2, 67.2)  & (96.2, 93.8, 66.8)  \\
        \bottomrule 
    \end{tabular}
    }
\end{table*}
\begin{table*}[!bp]
    \centering
    \caption{\textbf{Category-level anomaly classification performance for the dataset DAGM.} each triplet reports (AUROC, AP, F1-max).}
    \vspace{\TableCaptionGap}
    \small
    \resizebox{1\linewidth}{!}{
    \begin{tabular}{lccccccc}
        \toprule
        Product & AnVoL & VAND & AnomalyCLIP & AdaCLIP & \textbf{\methodname} & \textbf{\methodnamep} \\ \midrule
        \midrule
        Class01  & (58.3, 22.8, 30.8)  & (79.9, 37.0, 38.2)  & (85.8, 49.8, 52.6)  & (88.3, 44.2, 51.5)  & (92.8, 78.8, 73.2)  & (89.9, 66.8, 65.2)  \\
        Class02  & (99.4, 98.1, 95.1)  & (95.0, 87.0, 81.3)  & (100 , 100 , 100 )  & (99.5, 98.5, 95.9)  & (100 , 100 , 100 )  & (100 , 100 , 100 )  \\
        Class03  & (99.6, 97.9, 93.0)  & (99.4, 96.6, 91.4)  & (99.9, 99.5, 97  )  & (100 , 100 , 100 )  & (100 , 100 , 100 )  & (100 , 100 , 100 )  \\
        Class04  & (89.9, 74.5, 54.3)  & (85.2, 66.4, 48.6)  & (97.6, 94.1, 85.7)  & (95.8, 89.2, 79.4)  & (99.4, 98.5, 96.4)  & (99.0, 98.1, 93.2)  \\
        Class05  & (95.7, 87.4, 72.8)  & (94.6, 82.2, 65.9)  & (99.2, 98.1, 100 )  & (98.9, 94.3, 90.2)  & (100 , 100 , 100 )  & (99.9, 99.6, 98.2)  \\
        Class06  & (99.1, 95.8, 87.3)  & (98.5, 92.4, 81.6)  & (99.8, 99.4, 100 )  & (99.9, 98.7, 96.1)  & (100 , 100 , 100 )  & (100 , 99.9, 99.1)  \\
        Class07  & (78.8, 54.6, 40.5)  & (86.1, 70.9, 58.3)  & (95.3, 90.5, 94.8)  & (94.7, 82.8, 73.6)  & (100 , 100 , 99.7)  & (98.3, 96.0, 90.7)  \\
        Class08  & (81.5, 60.3, 44.9)  & (90.3, 78.7, 67.2)  & (97.7, 95.2, 87.1)  & (96.5, 89.1, 83.1)  & (99.8, 99.0, 96.3)  & (99.5, 98.8, 96.7)  \\
        Class09  & (92.4, 83.2, 68.1)  & (94.8, 88.0, 75.5)  & (99.0, 98.3, 86.1)  & (98.7, 96.8, 92.5)  & (99.7, 98.3, 93.4)  & (99.9, 99.7, 98.4)  \\
        Class10  & (96.8, 90.5, 80.2)  & (92.0, 82.1, 70.7)  & (98.9, 97.5, 97.6)  & (98.5, 95.6, 89.8)  & (100 , 99.8, 99.0)  & (99.8, 99.3, 98.0)  \\
        \midrule
        \textbf{Mean} & (89.7, 76.1, 74.7)  & (94.4, 83.9, 79.9)  & (97.7, 92.4, 90.1)  & (96.9, 88.5, 87.7)  & (99.2, 97.4, 95.8)  & (98.9, 96.1, 94.7)  \\
        \bottomrule
    \end{tabular}
    }
\end{table*}

\begin{table*}[!bp]
    \centering
    \caption{\textbf{Category-level anomaly localization performance for the dataset DTD-Syn.} each triplet reports (AUROC, AUPRO, F1-max).}
    \vspace{\TableCaptionGap}
    \small
    \resizebox{1\linewidth}{!}{
        \begin{tabular}{lccccccc}
            \toprule
            Product & AnVoL & VAND & AnomalyCLIP & AdaCLIP & \textbf{\methodname} & \textbf{\methodnamep} \\ \midrule
            \midrule
            Woven\_001      & (93.0, 75.6, 33.8)  & (99.2, 82.6, 77.9)  & (99.7, 98.9, 67.2)  & (99.9, 87.1, 78.0)  & (99.8, 99.1, 72.7)  & (99.9, 99.4, 75.0)  \\
            Woven\_127      & (89.4, 74.9, 19.1)  & (90.8, 55.6, 60.2)  & (93.7, 89.5, 51.9)  & (96.0, 65.3, 64.2)  & (95.6, 94.7, 63.6)  & (95.3, 93.7, 65.9)  \\
            Woven\_104      & (96.1, 86.5, 41.5)  & (94.3, 69.5, 68.9)  & (96.1, 92.5, 63.1)  & (98.6, 79.4, 73.1)  & (98.4, 96.7, 68.7)  & (98.6, 96.8, 70.9)  \\
            Stratified\_154 & (99.2, 94.6, 61.1)  & (96.8, 77.6, 78.6)  & (99.5, 96.2, 67.4)  & (97.5, 76.5, 72.4)  & (99.5, 99.0, 72.3)  & (99.3, 98.3, 72.9)  \\
            Blotchy\_099    & (94.4, 84.1, 37.3)  & (99.0, 71.0, 68.5)  & (99.5, 96.2, 67.5)  & (99.7, 87.3, 79.2)  & (99.6, 96.5, 73.2)  & (99.7, 97.2, 76.0)  \\
            Woven\_068      & (97.2, 89.1, 33.1)  & (95.2, 63.4, 62.9)  & (98.7, 92.8, 47.8)  & (98.4, 65.1, 60.7)  & (98.7, 95.7, 51.6)  & (99.1, 96.9, 61.4)  \\
            Woven\_125      & (90.4, 80.8, 33.8)  & (98.8, 84.6, 83.5)  & (99.4, 95.6, 64.1)  & (99.8, 90.6, 82.5)  & (99.6, 97.9, 70.5)  & (99.7, 99.1, 75.5)  \\
            Marbled\_078    & (97.7, 92  , 43.6)  & (98.1, 77.4, 73.3)  & (99.1, 97.1, 62  )  & (99.6, 85.2, 77.1)  & (99.4, 97.7, 68.3)  & (99.4, 97.6, 71.3)  \\
            Perforated\_037 & (98.8, 95.9, 46.3)  & (89.0, 61.0, 68.1)  & (94.6, 85.1, 63.1)  & (96.4, 70.6, 69.2)  & (96.6, 94.1, 70.0)  & (97.9, 96.5, 71.7)  \\
            Mesh\_114       & (83.4, 57.7, 26.4)  & (89.0, 60.6, 66.4)  & (95.2, 77.0, 56.5)  & (97.7, 73.7, 70.0)  & (95.2, 83.3, 64.8)  & (97.0, 88.7, 68.1)  \\
            Fibrous\_183    & (93.8, 80.2, 35.6)  & (97.5, 56.1, 55.7)  & (99.4, 98.2, 69.2)  & (99.0, 82.1, 75.0)  & (99.7, 99.2, 78.6)  & (99.7, 99.1, 77.1)  \\
            Matted\_069     & (86.1, 61.6, 17.5)  & (95.2, 44.1, 45.1)  & (99.6, 84.8, 66.7)  & (98.5, 56.4, 55.5)  & (99.7, 88.4, 74.5)  & (99.7, 89.2, 75.9)  \\
            \midrule
            \textbf{Mean}   & (93.3, 81.1, 35.8)  & (95.2, 66.9, 67.4)  & (97.9, 92.0, 62.2)  & (98.4, 76.6, 71.4)  & (98.5, 95.2, 69.1)  & (98.8, 96.0, 71.8)  \\
            \bottomrule
        \end{tabular}
    }
\end{table*}
\begin{table*}[!htbp]
    \centering
    \caption{\textbf{Category-level anomaly classification performance for the dataset DTD-Syn.} each triplet reports (AUROC, AP, F1-max).}
    \vspace{\TableCaptionGap}
    \small
    \resizebox{1\linewidth}{!}{
        \begin{tabular}{lccccccc}
            \toprule
            Product & AnVoL & VAND & AnomalyCLIP & AdaCLIP & \textbf{\methodname} & \textbf{\methodnamep} \\ \midrule
            \midrule
            Woven\_001      & (96.7, 94.0, 98.7) & (96.1, 95.5, 98.6)  & (100,  100 , 100 )  & (100 , 100 , 100 )  & (99.4, 99.9, 98.7)  & (98.4, 99.6, 97.5)  \\
            Woven\_127      & (80.8, 77.9, 83.8) & (74.4, 70.2, 77.8)  & (80.7, 83.5, 76.2)  & (99.8, 99.9, 98.8)  & (100 , 100 , 100 )  & (98.6, 99.7, 98.1)  \\
            Woven\_104      & (94.8, 93.9, 98.7) & (76.2, 89.9, 93.7)  & (98.1, 99.6, 97.5)  & (97.9, 99.4, 98.2)  & (98.6, 99.7, 98.7)  & (100 , 100 , 99.3)  \\
            Stratified\_154 & (99.8, 98.8, 99.9) & (97.4, 96.3, 99.4)  & (97.6, 99.4, 95.8)  & (91.7, 97.7, 95.1)  & (99.7, 99.9, 98.8)  & (99.8, 100 , 99.4)  \\
            Blotchy\_099    & (99.3, 98.7, 99.8) & (92.6, 92.0, 98.2)  & (98.9, 99.7, 98.8)  & (89.6, 95.4, 90.5)  & (92.6, 94.3, 87.4)  & (90.5, 92.6, 83.3)  \\
            Woven\_068      & (86.2, 81.7, 92.6) & (84.5, 80.0, 91.6)  & (96.9, 98.4, 94.9)  & (92.6, 98.2, 93.1)  & (99.0, 99.8, 98.8)  & (100 , 100 , 100 )  \\
            Woven\_125      & (99.7, 99.4, 99.9) & (94.3, 93.9, 98.5)  & (99.8, 100 , 99.4)  & (99.4, 99.9, 98.1)  & (100 , 100 , 100 )  & (98.9, 99.8, 98.8)  \\
            Marbled\_078    & (98.7, 98.1, 99.7) & (98.8, 98.1, 99.7)  & (98.7, 99.7, 97.5)  & (99.1, 99.6, 97.8)  & (93.7, 98.5, 93.3)  & (95.2, 98.9, 94.5)  \\
            Perforated\_037 & (99.9, 99.4, 100 ) & (75.1, 88.9, 92.9)  & (90.6, 97.5, 92.5)  & (89.5, 94.0, 84.7)  & (87.6, 95.1, 86.6)  & (88.3, 95.2, 85.9)  \\
            Mesh\_114       & (88.2, 86.1, 95.4) & (72.7, 81.7, 87.7)  & (85.8, 94.5, 84.4)  & (99.3, 99.8, 99.4)  & (95.1, 97.3, 91.0)  & (94.8, 97.2, 90.5)  \\
            Fibrous\_183    & (98.1, 97.5, 99.6) & (89.4, 92.8, 97.2)  & (97.2, 99.3, 95.7)  & (100 , 100 , 100 )  & (99.1, 99.8, 97.5)  & (97.3, 99.4, 96.2)  \\
            Matted\_069     & (96.6, 94.6, 99.2) & (74.7, 88.8, 92.5)  & (82.6, 95.2, 91.2)  & (85.1, 83.5, 81.8)  & (91.1, 97.8, 92.2)  & (87.8, 96.8, 91.8)  \\
            \midrule
            \textbf{Mean}   & (94.9, 93.3, 97.3) & (85.5, 89.0, 94.0)  & (93.9, 97.2, 93.6)  & (95.3, 97.3, 94.8)  & (96.3, 98.5, 95.3)  & (95.8, 98.3, 94.6)  \\
            \bottomrule
        \end{tabular}
    }
\end{table*}

\begin{table*}[!htbp]
    \centering
    \caption{\textbf{Category-level anomaly localization performance for the dataset MPDD.} each triplet reports (AUROC, AUPRO, F1-max).}
    \vspace{\TableCaptionGap}
    \small
    \resizebox{1\linewidth}{!}{
        \begin{tabular}{lccccccc}
            \toprule
            Product & AnVoL & VAND & AnomalyCLIP & AdaCLIP & \textbf{\methodname} & \textbf{\methodnamep} \\ \midrule
            \midrule
            bracket\_black & (24.6, 1.6 , 0.2 ) & (96.3, 90.6, 15.8)  & (95.7, 85.2, 27.2)  & (96.5, 82.6, 14.4)  & (96.1, 86.1, 28.6)  & (97.2, 90.4, 30.4)  \\
            bracket\_brown & (27.4, 3.4 , 1.0 ) & (87.4, 72.6, 8.7 )  & (94.4, 77.8, 13.1)  & (93.3, 28.7, 18.8)  & (99.8, 96.2, 35.1)  & (99.8, 97.7, 31.4)  \\
            bracket\_white & (45.2, 1.7 , 0.1 ) & (99.2, 93.7, 8.9 )  & (99.8, 98.8, 22.9)  & (98.1, 63.9, 6.9 )  & (97.7, 92.2, 25.2)  & (98.0, 95.5, 36.6)  \\
            connector      & (90.0, 68.2, 4.7 ) & (90.6, 74.5, 22.5)  & (97.2, 89.9, 27.0)  & (97.4, 77.9, 39.2)  & (95.6, 82.1, 17.2)  & (96.1, 89.1, 16.2)  \\
            metal\_plate   & (95.9, 85.3, 70.4) & (93.0, 74.5, 63.1)  & (93.7, 86.8, 61.9)  & (92.0, 33.5, 57.8)  & (93.9, 83.4, 62.5)  & (95.3, 89.3, 66.8)  \\
            tubes          & (90.7, 69.7, 17.3) & (99.1, 96.9, 68.7)  & (98.1, 93.6, 53.3)  & (98.9, 90.4, 70.0)  & (98.9, 95.7, 61.0)  & (99.2, 97.1, 70.2)  \\
            \midrule
            \textbf{Mean}  & (62.3, 38.3, 15.6) & (94.3, 83.8, 31.3)  & (96.5, 88.7, 34.2)  & (96.0, 62.8, 34.5)  & (97.0, 89.3, 38.2)  & (97.6, 93.2, 41.9)  \\
            \bottomrule
        \end{tabular}
    }
\end{table*}

\begin{table*}[!htbp]
    \centering
    \caption{\textbf{Category-level anomaly classification performance for the dataset MPDD.} each triplet reports (AUROC, AP, F1-max).}
    \vspace{\TableCaptionGap}
    \small
    \resizebox{1\linewidth}{!}{
        \begin{tabular}{lccccccc}
            \toprule
            Product & AnVoL & VAND & AnomalyCLIP & AdaCLIP & \textbf{\methodname} & \textbf{\methodnamep} \\ \midrule
            \midrule
            bracket\_black & (42.5,  59.7,  76.5) & (68.4, 72.6, 80.0)  & (67.8, 73.4, 78.6)  & (71.4, 81.1, 77.7)  & (82.1, 74.1, 66.7)  & (83.3, 70.3, 71.8)  \\
            bracket\_brown & (66.7,  92.3,  90.7) & (61.6, 78.0, 81.0)  & (62.0, 80.4, 80.3)  & (51.9, 71.8, 79.7)  & (80.8, 81.8, 76.9)  & (84.0, 82.7, 82.5)  \\
            bracket\_white & (38.5,  58.2,  76.5) & (85.7, 88.2, 78.1)  & (67.7, 71.6, 69.8)  & (77.8, 80.0, 74.7)  & (72.9, 77.7, 81.1)  & (76.9, 81.8, 79.6)  \\
            connector      & (100  , 100  , 100 ) & (78.5, 71.7, 66.7)  & (87.4, 77.0, 73.7)  & (64.4, 61.9, 58.3)  & (64.9, 80.1, 81.0)  & (54.6, 74.4, 79.7)  \\
            metal\_plate   & (98.3,  99.4,  95.9) & (69.9, 86.5, 86.6)  & (84.7, 94.4, 87.5)  & (86.6, 94.8, 90.3)  & (91.4, 96.8, 90.9)  & (89.9, 96.6, 89.7)  \\
            tubes          & (90.4,  95.4,  90.4) & (95.7, 98.1, 92.2)  & (95.4, 98.1, 92.3)  & (80.8, 91.6, 84.7)  & (96.4, 98.5, 93.0)  & (97.5, 99.0, 94.8)  \\
            \midrule
            \textbf{Mean}  & (72.7,  83.6,  88.3) & (76.6, 82.5, 80.8)  & (77.5, 82.5, 80.4)  & (72.1, 80.2, 77.6)  & (81.4, 84.8, 81.6)  & (81.0, 84.1, 83.0)  \\
            \bottomrule
        \end{tabular}
    }
\end{table*}

\begin{table*}[!htbp]
    \centering
    \caption{\textbf{Category-level anomaly localization performance for the dataset MVTec-AD.} each triplet reports (AUROC, AUPRO, F1-max).}
    \vspace{\TableCaptionGap}
    \small
    \resizebox{1\linewidth}{!}{
    \begin{tabular}{lccccccc}
        \toprule
        Product & AnVoL & VAND & AnomalyCLIP & AdaCLIP & \textbf{\methodname} & \textbf{\methodnamep} \\ \midrule
        \midrule
        bottle     & (90.9, 75.7, 53.2) & (83.5, 45.6, 53.4)  & (90.4, 80.8, 51.6)  & (90.8, 57.6, 60.8)  & (91.6, 84.6, 54.6)  & (93.6, 88.2, 60.6)  \\
        grid       & (67.7, 66.3, 23.0) & (72.3, 25.7, 23.9)  & (78.9, 64.0, 18.9)  & (78.3, 35.3, 26.5)  & (97.7, 80.0, 35.7)  & (99.2, 95.5, 48.1)  \\
        carpet     & (82.8, 52.4, 14.9) & (92.0, 51.3, 33.1)  & (95.8, 87.6, 31.0)  & (95.2, 18.0, 32.9)  & (99.0, 91.9, 59.2)  & (99.3, 95.8, 66.8)  \\
        capsule    & (95.6, 85.1, 42.5) & (98.4, 48.5, 65.7)  & (98.8, 90.0, 57.0)  & (98.9, 36.1, 67.4)  & (96.0, 90.5, 33.8)  & (95.5, 92.4, 33.5)  \\
        cable      & (95.7, 87.1, 22.3) & (95.8, 31.6, 40.8)  & (97.3, 75.4, 32.0)  & (97.0, 20.3, 39.0)  & (77.7, 68.8, 23.1)  & (79.4, 71.5, 22.0)  \\
        hazelnut   & (93.8, 75.6, 32.5) & (96.1, 70.2, 50.5)  & (97.2, 92.5, 47.6)  & (96.5, 59.2, 40.1)  & (97.1, 94.3, 50.6)  & (97.4, 94.8, 53.2)  \\
        leather    & (98.7, 94.4, 36.3) & (99.1, 72.4, 50.0)  & (98.6, 92.2, 33.2)  & (99.3, 76.9, 47.7)  & (99.1, 97.2, 40.7)  & (99.1, 98.3, 41.4)  \\
        metal\_nut & (71.4, 46.5, 29.6) & (65.5, 38.4, 28.0)  & (74.6, 71.1, 33.1)  & (74.4, 62.4, 35.3)  & (72.7, 74.4, 33.4)  & (76.9, 80.1, 34.0)  \\
        screw      & (79.5, 69.8, 18.4) & (76.2, 65.4, 27.7)  & (91.8, 88.1, 35.5)  & (87.7, 27.9, 35.7)  & (98.4, 91.7, 33.8)  & (98.6, 92.8, 39.3)  \\
        pill       & (88.5, 60.1, 13.5) & (97.8, 67.1, 41.7)  & (97.5, 88.0, 33.4)  & (98.3, 70.3, 34.5)  & (89.1, 90.8, 28.7)  & (85.1, 91.3, 27.5)  \\
        toothbrush & (77.0, 54.3, 35.9) & (92.7, 26.7, 66.5)  & (94.7, 87.4, 64.9)  & (91.1, 30.1, 61.9)  & (93.5, 90.4, 35.8)  & (89.7, 90.8, 29.5)  \\
        wood       & (91.6, 80.1, 20.6) & (95.8, 54.5, 48.1)  & (91.9, 88.5, 29.0)  & (94.7, 69.4, 37.9)  & (97.5, 93.8, 60.7)  & (96.7, 96.0, 60.8)  \\
        transistor & (75.6, 50.9, 24.6) & (62.4, 21.3, 19.0)  & (70.8, 58.2, 18.8)  & (57.8, 31.2, 16.3)  & (69.5, 56.0, 17.7)  & (65.4, 54.8, 16.7)  \\
        tile       & (95.1, 75.1, 49.7) & (95.8, 31.1, 60.3)  & (96.4, 91.5, 55.2)  & (92.6, 48.1, 56.0)  & (96.1, 87.5, 65.7)  & (95.5, 90.3, 68.0)  \\
        zipper     & (94.3, 82.2, 35.1) & (91.1, 10.7, 40.5)  & (91.3, 65.4, 45.0)  & (95.8, 18.2, 57.2)  & (95.1, 77.3, 46.3)  & (97.3, 89.3, 54.9)  \\
        \midrule
        \textbf{Mean}       & (86.6, 70.4, 30.1) & (87.6, 44.0, 43.3)  & (91.1, 81.4, 39.1)  & (89.9, 44.1, 43.4)  & (91.3, 84.6, 41.3)  & (91.2, 88.1, 43.8)  \\
        \bottomrule
    \end{tabular}
    }
\end{table*}

\begin{table*}[!htbp]
    \centering
    \caption{\textbf{Category-level anomaly classification performance for the dataset MVTec-AD.} each triplet reports (AUROC, AP, F1-max).}
    \vspace{\TableCaptionGap}
    \small
    \resizebox{1\linewidth}{!}{
    \begin{tabular}{lccccccc}
        \toprule
        Product & AnVoL & VAND & AnomalyCLIP & AdaCLIP & \textbf{\methodname} & \textbf{\methodnamep} \\ \midrule
        \midrule
        bottle     & (96.3, 99.0, 96.8) & (91.8, 97.6, 92.1)  & (88.7, 96.8, 90.9)  & (97.8, 99.3, 95.4)  & (91.3, 97.5, 91.1)  & (92.4, 97.7, 92.8)  \\
        grid       & (87.8, 92.0, 86.0) & (88.3, 93.0, 85.1)  & (70.3, 81.7, 77.4)  & (64.3, 79.2, 76.0)  & (99.7, 99.9, 98.3)  & (100 , 100 , 100 )  \\
        carpet     & (79.1, 94.5, 92.7) & (79.8, 95.4, 92.0)  & (89.5, 97.8, 91.7)  & (84.6, 96.6, 92.0)  & (100 , 100 , 100 )  & (99.9, 100 , 99.4)  \\
        capsule    & (99.3, 99.8, 99.4) & (99.5, 99.8, 98.3)  & (100 , 100 , 99.4)  & (100 , 100 , 100 )  & (93.5, 98.6, 94.1)  & (92.3, 98.3, 93.8)  \\
        cable      & (98.1, 99.4, 96.4) & (86.4, 95.0, 89.1)  & (97.8, 99.3, 97.3)  & (97.7, 99.1, 97.4)  & (87.1, 92.5, 83.9)  & (88.1, 93.4, 86.2)  \\
        hazelnut   & (94.0, 96.9, 89.4) & (89.5, 94.7, 87.0)  & (97.2, 98.5, 92.6)  & (87.0, 93.0, 86.1)  & (98.5, 99.3, 96.4)  & (98.0, 99.1, 96.3)  \\
        leather    & (100 , 100 , 100 ) & (99.7, 99.9, 98.9)  & (99.8, 99.9, 99.5)  & (99.9, 99.9, 99.5)  & (100 , 100 , 100 )  & (100 , 100 , 100 )  \\
        metal\_nut & (97.4, 99.4, 96.7) & (68.5, 91.9, 89.4)  & (92.4, 98.2, 93.7)  & (66.6, 92.1, 89.4)  & (79.9, 95.5, 90.3)  & (84.1, 96.4, 89.9)  \\
        screw      & (86.6, 97.2, 91.8) & (80.6, 96.0, 91.6)  & (81.1, 95.3, 92.1)  & (88.9, 97.6, 94.0)  & (90.0, 95.8, 91.6)  & (92.4, 97.3, 92.1)  \\
        pill       & (78.0, 92.1, 87.1) & (84.7, 93.5, 88.8)  & (82.1, 92.9, 88.3)  & (88.1, 95.0, 90.0)  & (84.8, 96.7, 93.5)  & (87.0, 97.5, 92.2)  \\
        toothbrush & (100 , 100 , 99.4) & (99.9, 99.9, 98.8)  & (100 , 100 , 100 )  & (100 , 100 , 100 )  & (95.8, 98.7, 95.1)  & (86.4, 95.1, 88.5)  \\
        wood       & (92.2, 96.9, 93.1) & (54.0, 72.2, 83.3)  & (85.3, 93.9, 90.0)  & (91.1, 97.0, 90.9)  & (98.6, 99.6, 97.4)  & (97.9, 99.4, 97.5)  \\
        transistor & (86.4, 86.5, 77.9) & (81.1, 77.6, 74.5)  & (93.9, 92.1, 83.7)  & (86.8, 87.4, 78.7)  & (90.6, 88.6, 79.1)  & (91.8, 90.2, 79.5)  \\
        tile       & (99.3, 99.8, 98.3) & (99.0, 99.7, 96.8)  & (96.9, 99.2, 96.6)  & (99.0, 99.7, 96.7)  & (99.7, 99.9, 98.8)  & (99.7, 99.9, 98.8)  \\
        zipper     & (92.9, 97.9, 93.3) & (89.5, 97.1, 90.8)  & (98.4, 99.5, 97.9)  & (99.3, 99.8, 98.3)  & (97.6, 99.3, 97.9)  & (99.0, 99.7, 97.5)  \\ 
        \midrule
        \textbf{Mean}       & (92.5, 96.7, 93.2) & (86.2, 93.6, 90.4)  & (91.6,  96.4, 92.7)  & (90.1,  95.7, 92.3)  & (93.8, 97.5, 93.8)  & (93.9, 97.6, 93.6)  \\ \bottomrule
    \end{tabular}
    }
\end{table*}

\begin{table*}[!htbp]
    \centering
    \caption{\textbf{Category-level anomaly localization performance for the dataset ViSA.} each triplet reports (AUROC, AUPRO, F1-max).}
    \vspace{\TableCaptionGap}
    \small
    \resizebox{1\linewidth}{!}{
        \begin{tabular}{lccccccc}
            \toprule
            Product & AnVoL & VAND & AnomalyCLIP & AdaCLIP & \textbf{\methodname} & \textbf{\methodnamep} \\ \midrule
            \midrule
            candle      & (95.6, 83.4, 14.7) & (97.8, 92.5, 39.4) & (80.9, 82.6, 37.8) & (99.2, 76.7, 48.2)  & (86.2, 92.2, 16.7)  & (90.4, 94.9, 13.5)  \\
            capsules    & (82.9, 44.4, 9.8 ) & (97.5, 86.7, 48.5) & (82.8, 89.4, 37.8) & (98.7, 76.1, 47.6)  & (99.0, 87.3, 67.1)  & (95.7, 89.3, 32.6)  \\
            cashew      & (89.8, 85.7, 11.1) & (86.0, 91.7, 22.9) & (76.0, 89.3, 25.8) & (95.9, 49.3, 34.2)  & (96.8, 88.4, 43.4)  & (99.4, 93.7, 69.6)  \\
            chewinggum  & (91.4, 56.0, 49.3) & (99.5, 87.2, 78.5) & (97.2, 98.8, 61.0) & (99.5, 64.9, 79.5)  & (94.8, 84.7, 33.7)  & (93.8, 90.7, 27.3)  \\
            fryum       & (77.9, 56.4, 23.8) & (92.0, 89.7, 29.7) & (92.7, 96.6, 30.3) & (94.1, 69.2, 28.9)  & (98.5, 92.3, 37.6)  & (98.9, 97.5, 40.3)  \\
            macaroni1   & (81.9, 41.0, 1.1 ) & (98.8, 93.2, 35.5) & (86.7, 85.5, 23.7) & (99.7, 82.8, 35.0)  & (92.0, 86.8, 15.7)  & (93.3, 89.5, 17.5)  \\
            macaroni2   & (78.0, 34.4, 0.1 ) & (97.8, 82.3, 13.7) & (72.2, 70.8, 5.1 ) & (99.0, 72.1, 14.2)  & (93.7, 81.2, 22.3)  & (93.9, 83.7, 25.0)  \\
            pcb1        & (91.1, 72.1, 17.9) & (92.7, 87.5, 12.5) & (85.2, 86.7, 12.7) & (92.5, 59.0, 23.8)  & (89.8, 79.5, 15.6)  & (88.4, 79.6, 16.1)  \\
            pcb2        & (85.1, 54.4, 3.0 ) & (89.7, 75.5, 23.4) & (62.0, 64.4, 15.8) & (92.3, 78.8, 30.6)  & (98.9, 90.0, 29.2)  & (98.0, 86.0, 7.4 )  \\
            pcb3        & (76.0, 32.8, 1.2 ) & (88.4, 77.8, 21.7) & (61.7, 69.4, 9.3 ) & (87.9, 77.3, 33.7)  & (98.3, 86.0, 07.3)  & (99.2, 93.0, 34.2)  \\
            pcb4        & (93.3, 78.4, 33.0) & (94.6, 86.8, 31.3) & (93.9, 94.3, 34.7) & (96.3, 87.5, 43.6)  & (98.0, 91.5, 43.2)  & (97.4, 97.4, 37.8)  \\
            pipe\_fryum & (79.2, 87.5, 10.8) & (96.0, 90.9, 30.4) & (92.3, 96.3, 45.5) & (97.4, 81.7, 36.0)  & (94.9, 90.0, 39.2)  & (95.2, 91.6, 40.9)  \\
            \midrule
            \textbf{Mean}        & (85.2, 60.5, 14.6) & (94.2, 86.8, 32.3)  & (95.5, 86.7, 28.3)  & (96.0, 72.9, 37.9)  & (95.1, 87.5, 30.9)  & (95.3, 90.6, 30.2)  \\
            \bottomrule
        \end{tabular}
    }
\end{table*}

\begin{table*}[!htbp]
    \centering
    \caption{\textbf{Category-level anomaly classification performance for the dataset ViSA.} each triplet reports (AUROC, AP, F1-max).}
    \vspace{\TableCaptionGap}
    \small
    \resizebox{1\linewidth}{!}{
        \begin{tabular}{lccccccc}
            \toprule
            Product & AnVoL & VAND & AnomalyCLIP & AdaCLIP & \textbf{\methodname} & \textbf{\methodnamep} \\ \midrule
            \midrule
            candle      & (97.2, 97.2, 92.5) & (83.5, 86.6, 77.1)  & (98.8, 96.5, 75.6)  & (94.5, 95.8, 89.1)  & (83.3, 92.9, 82.2)  & (85.9, 93.7, 84.5)  \\
            capsules    & (80.6, 89.4, 80.5) & (61.4, 74.5, 78.0)  & (95.0, 78.9, 82.2)  & (74.3, 82.1, 80.9)  & (97.5, 99.0, 96.9)  & (81.5, 90.0, 80.2)  \\
            cashew      & (90.2, 95.7, 86.8) & (86.9, 94.0, 84.8)  & (93.8, 91.9, 80.3)  & (93.7, 97.3, 92.2)  & (84.4, 91.2, 83.7)  & (97.9, 99.2, 96.4)  \\
            chewinggum  & (96.7, 98.6, 94.8) & (96.5, 98.4, 93.7)  & (99.3, 90.9, 94.8)  & (91.6, 96.5, 89.6)  & (95.1, 97.8, 92.3)  & (92.7, 96.8, 90.2)  \\
            fryum       & (90.2, 95.6, 88.7) & (94.2, 97.2, 91.7)  & (94.6, 86.9, 90.1)  & (86.1, 93.5, 84.7)  & (87.1, 89.9, 80.0)  & (83.5, 87.2, 76.4)  \\
            macaroni1   & (70.4, 70.8, 71.4) & (71.4, 70.4, 71.7)  & (98.3, 89.8, 80.4)  & (73.2, 66.1, 75.7)  & (83.6, 84.9, 79.3)  & (83.5, 85.5, 77.0)  \\
            macaroni2   & (61.7, 61.2, 68.9) & (64.7, 63.3, 69.1)  & (97.6, 84.0, 71.2)  & (53.9, 52.8, 68.3)  & (74.0, 75.0, 72.5)  & (74.7, 76.5, 72.5)  \\
            pcb1        & (79.7, 82.4, 75.6) & (53.8, 57.4, 66.9)  & (94.0, 80.7, 78.8)  & (59.7, 64.1, 67.1)  & (68.2, 73.0, 66.4)  & (57.1, 65.1, 66.4)  \\
            pcb2        & (56.2, 53.9, 68.1) & (71.6, 73.8, 70.0)  & (92.4, 78.9, 67.8)  & (53.1, 56.6, 66.7)  & (85.4, 87.1, 77.6)  & (68.5, 66.4, 69.9)  \\
            pcb3        & (66.4, 68.2, 69.0) & (66.9, 70.8, 66.7)  & (88.4, 76.8, 66.4)  & (65.4, 69.8, 66.9)  & (69.1, 68.3, 71.0)  & (85.7, 86.5, 77.2)  \\
            pcb4        & (75.9, 75.7, 73.6) & (95.1, 95.2, 87.3)  & (95.7, 89.4, 87.8)  & (77.4, 79.9, 74.6)  & (98.7, 99.4, 96.0)  & (97.0, 98.5, 94.0)  \\
            pipe\_fryum & (85.4, 92.9, 85.5) & (89.7, 94.7, 87.8)  & (98.2, 96.2, 89.8)  & (85.9, 92.8, 86.1)  & (97.0, 96.8, 93.1)  & (94.9, 94.5, 89.2)  \\
            \midrule
            \textbf{Mean}  & (79.2, 81.6, 79.6) & (78.0, 81.4, 78.7)  & (82.0, 85.3, 80.4)  & (75.7, 79.0, 78.5)  & (85.3, 87.9, 82.6)  & (83.6, 86.6, 81.2)  \\
            \bottomrule
        \end{tabular}
    }
\end{table*}

\clearpage
\subsection{Category-level Qualitative Results}
\label{sec:category_level_qualitative_results}
To provide visual intuition of the method's capability in capturing anomalous patterns, we present zero-shot anomaly maps of \methodnamep across diverse domains, objects, and textures.
For MVTec-AD, we visualize results for the products capsule, carpet, grid, hazelnut, toothbrush, tile, and zipper. For ViSA, we illustrate anomaly maps for the categories 
candles, capsules, cashew, and pipe\_fryum. Localization predictions for the anomalous classes white\_brackets, tubes, and plates are reported for MPDD. In DTD-Synthetic, 
we provide visualizations for the Matted and Fibrous classes, while for DAGM, we include class06, class07, class08, and class09. Additionally, results for the single class 
from the KSDD dataset are presented. For medical anomaly detection, we provide zero-shot localization outputs for BrainMRI, ISIC, and CVC-ColonDB, using the medical training scheme 
discussed earlier at 
Appendix \ref{sec:ablation_medical_training}.

\begin{figure*}[!htbp]    
    \centering
    \includegraphics[width=\linewidth]{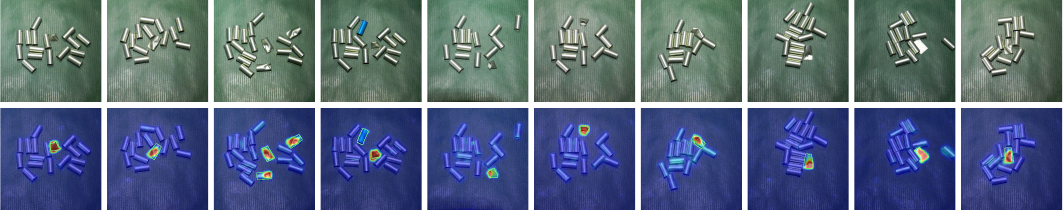}
    \vspace{\FigureCaptionGap}
    \caption{Localization score maps for the product, tubes, in MPDD dataset. The first row illustrates the original image, while the second row shows the anomaly segmentation results, with the regions encircled in green representing the ground truth.}
    \label{fig:appendix_tubes}
\end{figure*}

\mytextcolor{orange}{
\textit{Limitations.} Despite \methodname's strong performance and generalization capabilities, a performance gap remains compared to unsupervised methods that assume access to normal training samples and leverage full-shot training pipelines. This gap partly stems from the inherent challenge zero-shot models face in detecting unseen semantic anomalies, where domain knowledge is crucial to distinguish normal from anomalous samples. Figure~\ref{fig:appendix_tubes} illustrates this challenge: for instance, the blue tube is not flagged as anomalous because it lacks general structural defects, yet it is semantically inconsistent within the local context of the dataset.}


\begin{figure*}[!htbp]    
    \centering
    \includegraphics[width=\linewidth]{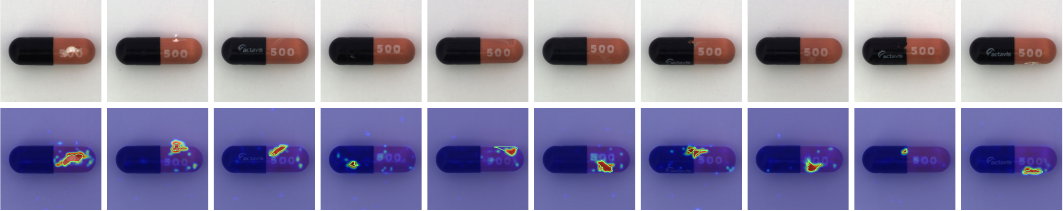}
    \vspace{\FigureCaptionGap}
    \caption{Localization score maps for the product, capsule, in MVTec-AD. The first row
    illustrates the original image, while the second row shows the anomaly segmentation
    results, with the regions encircled in green representing the ground truth.}
    \label{fig:appendix_capsule}
\end{figure*} 
\begin{figure*}[!htbp]    
    \centering
    \includegraphics[width=\linewidth]{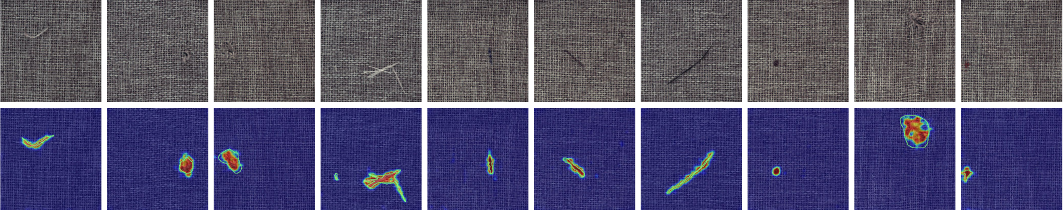}
    \vspace{\FigureCaptionGap}
    \caption{Localization score maps for the product, carpet, in MVTec-AD. The first row illustrates the original image, while the second row shows the anomaly segmentation results, with the regions encircled in green representing the ground truth.}
    \label{fig:appendix_carpet}
\end{figure*} 
\begin{figure*}[!htbp]    
    \centering
    \includegraphics[width=\linewidth]{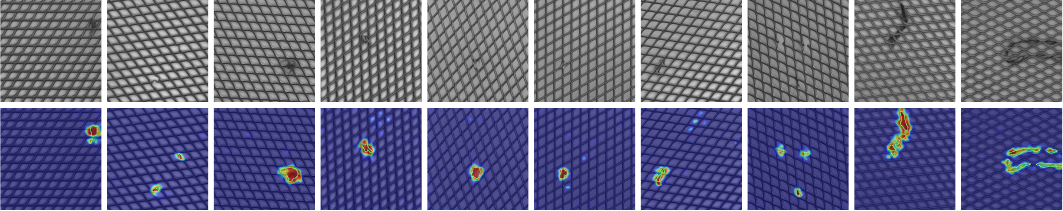}
    \vspace{\FigureCaptionGap}
    \caption{Localization score maps for the product, grid, in MVTec-AD. The first row illustrates the original image, while the second row shows the anomaly segmentation results, with the regions encircled in green representing the ground truth.}
    \label{fig:appendix_grid}
\end{figure*} 
\begin{figure*}[!htbp]    
    \centering
    \includegraphics[width=\linewidth]{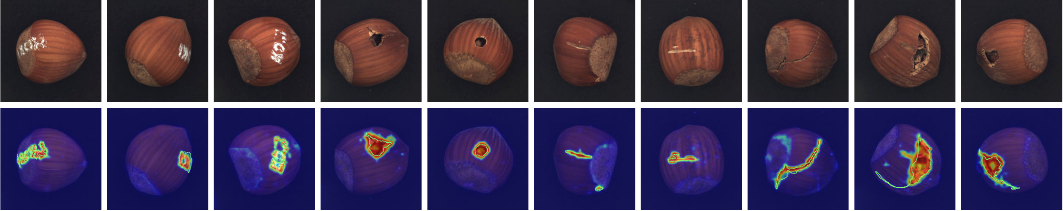}
    \vspace{\FigureCaptionGap}
    \caption{Localization score maps for the product, hazelnut, in MVTec-AD. The first row illustrates the original image, while the second row shows the anomaly segmentation results, with the regions encircled in green representing the ground truth.}
    \label{fig:appendix_hazelnut}
\end{figure*} 
\begin{figure*}[!htbp]    
    \centering
    \includegraphics[width=\linewidth]{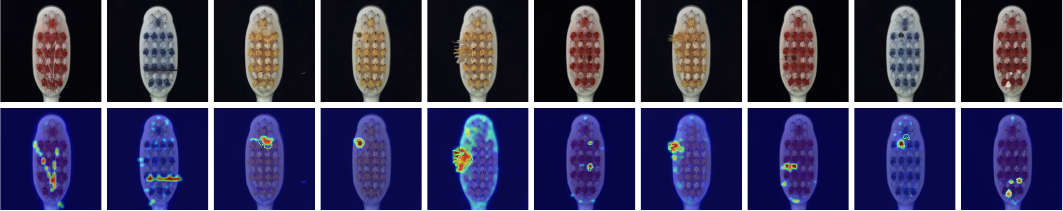}
    \vspace{\FigureCaptionGap}
    \caption{Localization score maps for the product, toothbrush, in MVTec-AD. The first row illustrates the original image, while the second row shows the anomaly segmentation results, with the regions encircled in green representing the ground truth.}
    \label{fig:appendix_toothbrush}
\end{figure*} 
\begin{figure*}[!htbp]    
    \centering
    \includegraphics[width=\linewidth]{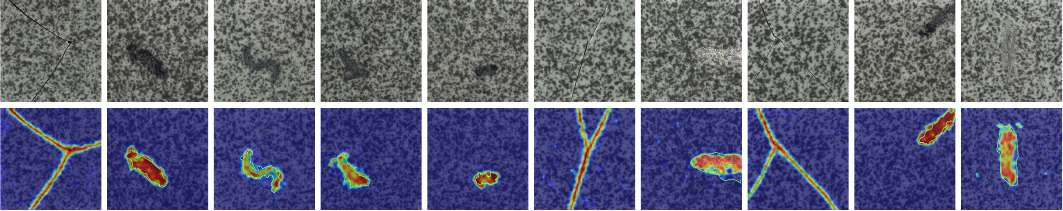}
    \vspace{\FigureCaptionGap}
    \caption{Localization score maps for the product, tile, in MVTec-AD. The first row illustrates the original image, while the second row shows the anomaly segmentation results, with the regions encircled in green representing the ground truth.}
    \label{fig:appendix_tile}
\end{figure*} 
\begin{figure*}[!htbp]    
    \centering
    \includegraphics[width=\linewidth]{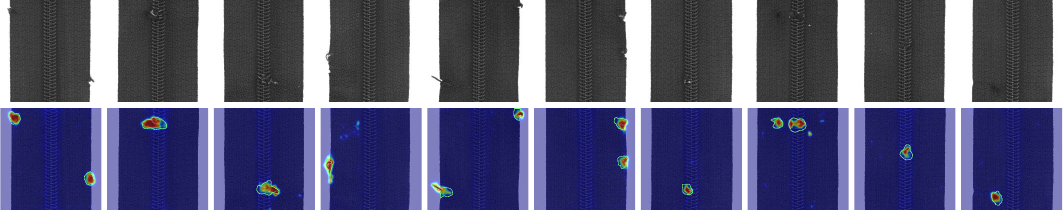}
    \vspace{\FigureCaptionGap}
    \caption{Localization score maps for the product, zipper, in MVTec-AD. The first row illustrates the original image, while the second row shows the anomaly segmentation results, with the regions encircled in green representing the ground truth.}
    \label{fig:appendix_zipper}
\end{figure*} 
\begin{figure*}[!htbp]    
    \centering
    \includegraphics[width=\linewidth]{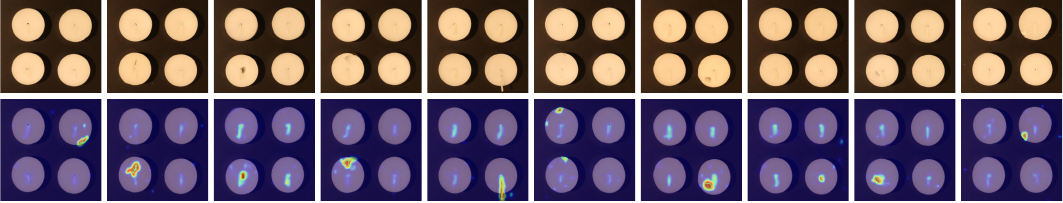}
    \vspace{\FigureCaptionGap}
    \caption{Localization score maps for the product, candles, in VISA dataset. The first row illustrates the original image, while the second row shows the anomaly segmentation results, with the regions encircled in green representing the ground truth.}
    \label{fig:appendix_candles}
\end{figure*} 
\begin{figure*}[!htbp]    
    \centering
    \includegraphics[width=\linewidth]{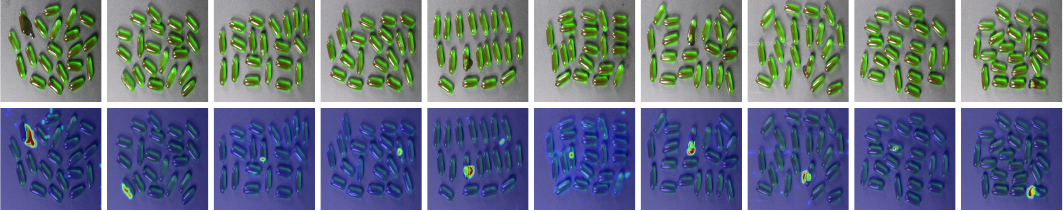}
    \vspace{\FigureCaptionGap}
    \caption{Localization score maps for the product, capsules, in VISA dataset. The first row illustrates the original image, while the second row shows the anomaly segmentation results, with the regions encircled in green representing the ground truth.}
    \label{fig:appendix_capsules}
\end{figure*} 
\begin{figure*}[!htbp]    
    \centering
    \includegraphics[width=\linewidth]{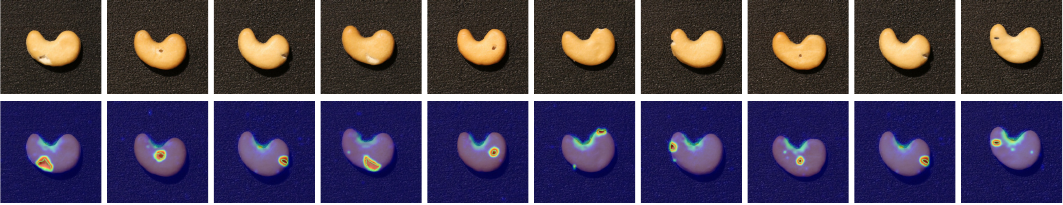}
    \vspace{\FigureCaptionGap}
    \caption{Localization score maps for the product, cashew, in VISA dataset. The first row illustrates the original image, while the second row shows the anomaly segmentation results, with the regions encircled in green representing the ground truth.}
    \label{fig:appendix_cashew}
\end{figure*} 
\begin{figure*}[!htbp]    
    \centering
    \includegraphics[width=\linewidth]{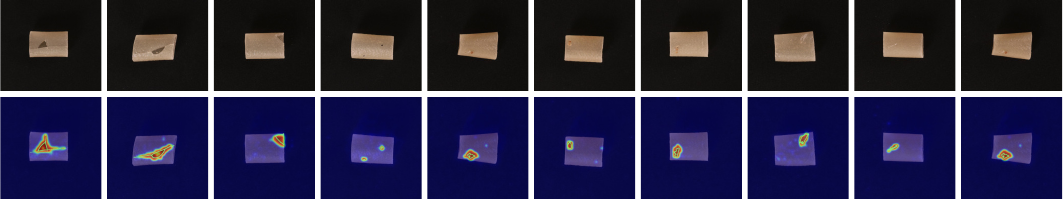}
    \vspace{\FigureCaptionGap}
    \caption{Localization score maps for the product, pipe fryum, in VISA dataset. The first row illustrates the original image, while the second row shows the anomaly segmentation results, with the regions encircled in green representing the ground truth.}
    \label{fig:appendix_pipe_fryom}
\end{figure*} 
\begin{figure*}[!htbp]    
    \centering
    \includegraphics[width=\linewidth]{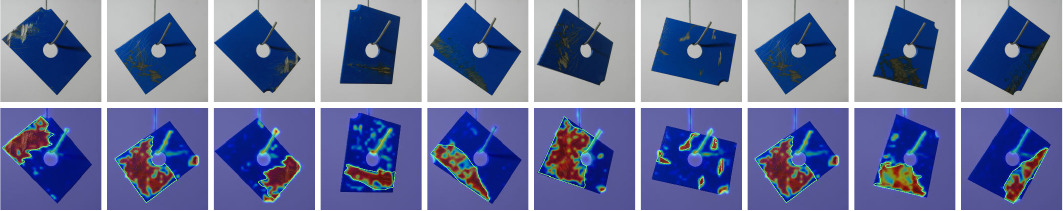}
    \vspace{\FigureCaptionGap}
    \caption{Localization score maps for the product, plates, in MPDD dataset. The first row illustrates the original image, while the second row shows the anomaly segmentation results, with the regions encircled in green representing the ground truth.}
    \label{fig:appendix_plates}
\end{figure*} 
\begin{figure*}[!htbp]    
    \centering
    \includegraphics[width=\linewidth]{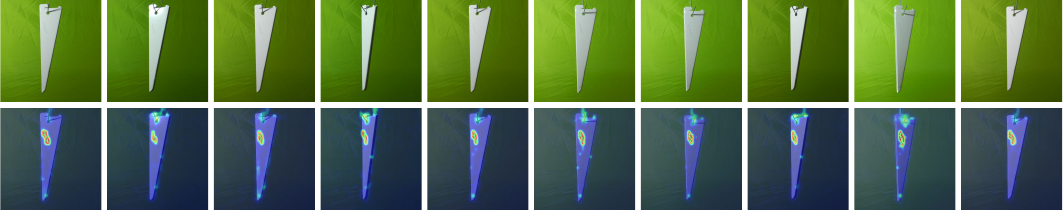}
    \vspace{\FigureCaptionGap}
    \caption{Localization score maps for the product, white brackets, in MPDD dataset. The first row illustrates the original image, while the second row shows the anomaly segmentation results, with the regions encircled in green representing the ground truth.}
    \label{fig:appendix_white_brackets}
\end{figure*} 
\begin{figure*}[!htbp]    
    \centering
    \includegraphics[width=\linewidth]{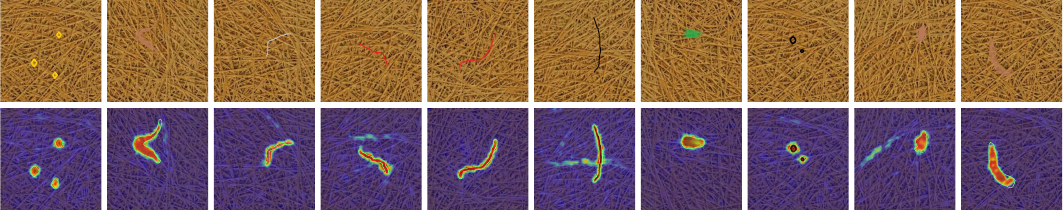}
    \vspace{\FigureCaptionGap}
    \caption{Localization score maps for the product, fibrous, in DTD dataset. The first row illustrates the original image, while the second row shows the anomaly segmentation results, with the regions encircled in green representing the ground truth.}
    \label{fig:appendix_fibrous}
\end{figure*} 
\begin{figure*}[!htbp]    
    \centering
    \includegraphics[width=\linewidth]{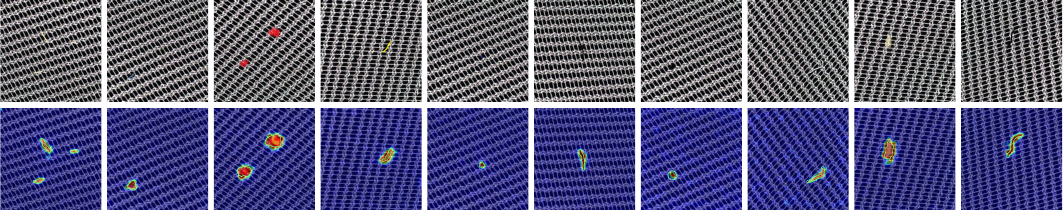}
    \vspace{\FigureCaptionGap}
    \caption{Localization score maps for the product, matted, in DTD dataset. The first row illustrates the original image, while the second row shows the anomaly segmentation results, with the regions encircled in green representing the ground truth.}
    \label{fig:appendix_matted}
\end{figure*} 
\begin{figure*}[!htbp]    
    \centering
    \includegraphics[width=\linewidth]{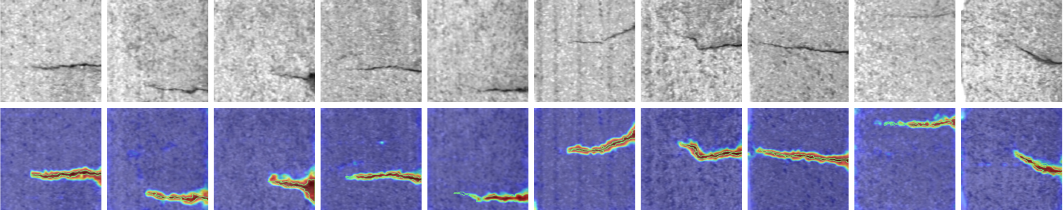}
    \vspace{\FigureCaptionGap}
    \caption{Localization score maps for the product, electric commutators, in SDD dataset. The first row illustrates the original image, while the second row shows the anomaly segmentation results, with the regions encircled in green representing the ground truth.}
    \label{fig:appendix_sdd}
\end{figure*} 
\begin{figure*}[!htbp]    
    \centering
    \includegraphics[width=\linewidth]{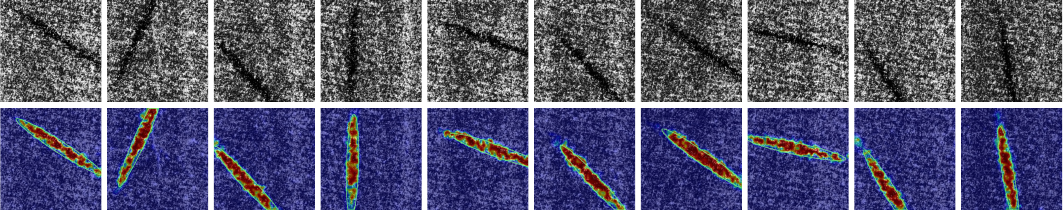}
    \vspace{\FigureCaptionGap}
    \caption{Localization score maps for the product, class 06, in DAGM dataset. The first row illustrates the original image, while the second row shows the anomaly segmentation results, with the regions encircled in green representing the ground truth.}
    \label{fig:appendix_class06}
\end{figure*} 
\begin{figure*}[!htbp]    
    \centering
    \includegraphics[width=\linewidth]{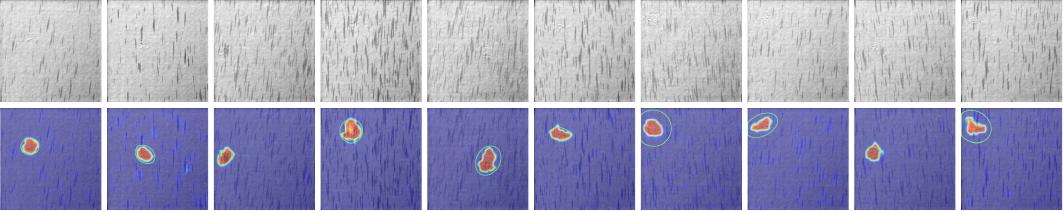}
    \vspace{\FigureCaptionGap}
    \caption{Localization score maps for the product, class 7, in DAGM dataset. The first row illustrates the original image, while the second row shows the anomaly segmentation results, with the regions encircled in green representing the ground truth.}
    \label{fig:appendix_class7}
\end{figure*} 
\begin{figure*}[!htbp]    
    \centering
    \includegraphics[width=\linewidth]{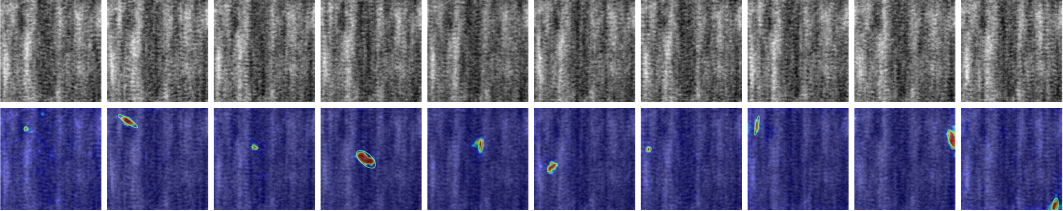}
    \vspace{\FigureCaptionGap}
    \caption{Localization score maps for the product, class 08, in DAGM dataset. The first row illustrates the original image, while the second row shows the anomaly segmentation results, with the regions encircled in green representing the ground truth.}
    \label{fig:appendix_class08}
\end{figure*} 
\begin{figure*}[!htbp]    
    \centering
    \includegraphics[width=\linewidth]{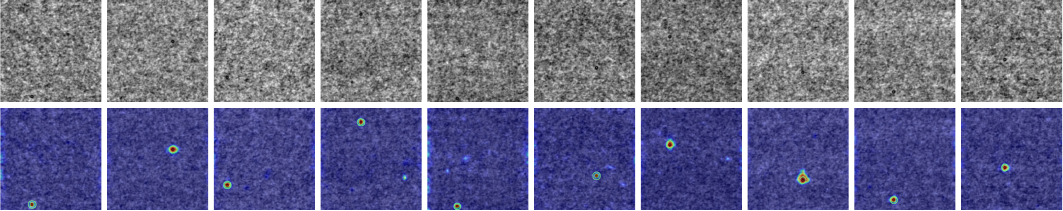}
    \vspace{\FigureCaptionGap}
    \caption{Localization score maps for the product, class 09, in DAGM dataset. The first row illustrates the original image, while the second row shows the anomaly segmentation results, with the regions encircled in green representing the ground truth.}
    \label{fig:appendix_class09}
\end{figure*} 
\begin{figure*}[!htbp]    
    \centering
    \includegraphics[width=\linewidth]{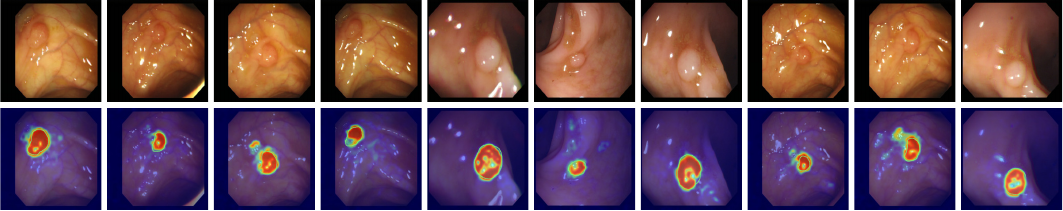}
    \vspace{\FigureCaptionGap}
    \caption{Localization score maps for the ColonDB dataset. The first row illustrates the original image, while the second row shows the anomaly segmentation results, with the regions encircled in green representing the ground truth.}
    \label{fig:appendix_colondb}
\end{figure*} 
\begin{figure*}[!htbp]    
    \centering
    \includegraphics[width=\linewidth]{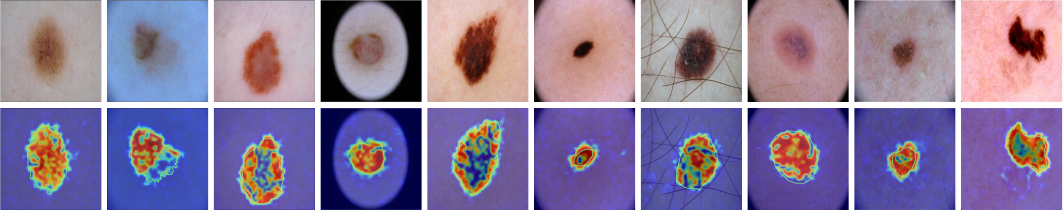}
    \vspace{\FigureCaptionGap}
    \caption{Localization score maps for the ISIC dataset. The first row illustrates the original image, while the second row shows the anomaly segmentation results, with the regions encircled in green representing the ground truth.}
    \label{fig:appendix_iciv}
\end{figure*} 
\begin{figure*}[!htbp]    
    \centering
    \includegraphics[width=\linewidth]{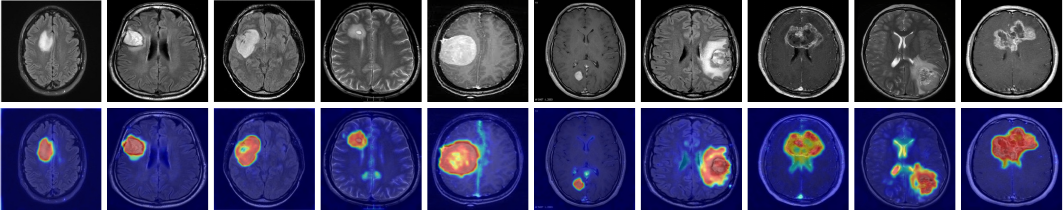}
    \vspace{\FigureCaptionGap}
    \caption{Localization score maps for the BrainMRI dataset. The first row illustrates the original image, while the second row shows the anomaly segmentation results, with the regions encircled in green representing the ground truth.}
    \label{fig:appendix_brainmri}
\end{figure*}

%% file: tables/ablations_increamental.tex
\begin{table}[h]
\centering
\caption{Illustrative incremental module ablation with AnomalyCLIP as the starting point. Performance is reported at pixel-level (AUROC, F1-max) and image-level (AUROC, F1-max).}
\vspace{\TableCaptionGap}
\small
\setlength{\tabcolsep}{5pt}
\begin{tabular}{lcccc}
\toprule
\multirow{2}{*}{\textbf{Module}} & \multicolumn{2}{c}{\textbf{MVTec-AD}} & \multicolumn{2}{c}{\textbf{VisA}} \\
 & \textbf{Pixel-level} & \textbf{Image-level} & \textbf{Pixel-level} & \textbf{Image-level} \\
\midrule
Baseline~(AnomalyCLIP) 
& (89.8, 36.9) & (91.0, 91.2) 
& (94.0, 25.2) & (80.1, 78.3) \\

+ Context-guided Prompt Learning 
& (90.6, 38.7) & (92.2, 92.3) 
& (94.5, 26.7) & (81.0, 79.2) \\

+ Extended Self-Correlation Attention 
& (91.2, 40.8) & (92.2, 92.3) 
& (94.3, 29.5) & (81.0, 79.2) \\

+ Score-based Aggregation 
& (91.2, 40.8) & (93.7, 93.9) 
& (94.3, 29.5) & (82.7, 80.4) \\

 \rowcolor{blue!10}+ D-Attn 
& (92.1, 44.7) & (94.7, 94.3) 
& (95.5, 29.2) & (82.6, 80.6) \\
\bottomrule
\end{tabular}

\label{tab:module_ablation}
\end{table}